\documentclass{article} 
\usepackage[final]{colm2026_conference}

\usepackage{microtype}
\usepackage{hyperref}
\usepackage{url}
\usepackage{booktabs}
\usepackage{graphicx}

\usepackage{algorithmicx}
\usepackage{algpseudocode}
\usepackage{algorithm}
\usepackage{subcaption}
\usepackage{amsmath}
\usepackage{amssymb}
\usepackage{mathtools}
\usepackage{amsthm}
\usepackage{xspace}
\usepackage{enumitem}
\usepackage{soul}
\usepackage{wrapfig}
\usepackage[capitalize,noabbrev]{cleveref}
\crefname{appendix}{Appendix}{Appendices}
\usepackage{multirow}
\usepackage{tcolorbox}

\setlist[itemize]{leftmargin=0.5cm}
\usepackage[
  separate-uncertainty = true,
  multi-part-units = repeat
]{siunitx}

\usepackage[textsize=tiny]{todonotes}
\tcbuselibrary{listings, skins, breakable}

\usepackage{lineno}

\definecolor{darkblue}{rgb}{0, 0, 0.5}
\hypersetup{colorlinks=true, citecolor=darkblue, linkcolor=darkblue, urlcolor=darkblue}

\title{Think Right: Learning to Mitigate Under-Over Thinking via Adaptive, Attentive Compression}

\author{\textbf{Joykirat Singh}\textsuperscript{1} \quad
\textbf{Justin Chih-Yao Chen}\textsuperscript{1} \quad
\textbf{Archiki Prasad}\textsuperscript{1} \\
\textbf{Elias Stengel-Eskin}\textsuperscript{2} \quad
\textbf{Akshay Nambi}\textsuperscript{3} \quad
\textbf{Mohit Bansal}\textsuperscript{1} \\
\textsuperscript{1}UNC Chapel Hill \quad
\textsuperscript{2}The University of Texas at Austin \quad
\textsuperscript{3}Microsoft Research
}

\newcommand{\shortname}{\texttt{TRAAC}\xspace}
\newcommand{\qwen}{Qwen3-4B\xspace}
\newcommand{\deepseek}{Deepseek-Qwen-7B\xspace}
\newcommand{\amc}{AMC\xspace}
\newcommand{\aime}{AIME\xspace}
\newcommand{\gpqa}{GPQA-D\xspace}
\newcommand{\underthinkBench}{UnderthinkingBench\xspace}
\newcommand{\overthinkBench}{OverthinkingBench\xspace}
\newcommand{\optimalThinkBench}{OptimalThinkingBench\xspace}
\newcommand{\bbeh}{BBEH\xspace}
\newcommand{\bbh}{BBH\xspace}

\begin{document}

\ifcolmsubmission
\linenumbers
\fi

\maketitle

\begin{abstract}
Recent thinking models are capable of solving complex reasoning tasks by scaling test-time compute, but this scaling should be allocated in line with task difficulty. On one hand, short reasoning (underthinking) leads to errors on harder problems that require extended reasoning steps; but, excessively long reasoning (overthinking) can be token-inefficient by generating unnecessary steps even after reaching a correct intermediate solution. We refer to this as \textbf{under-adaptivity}, where the model fails to modulate its response length appropriately given problems of varying difficulty. To address under-adaptivity and strike a balance between under- and overthinking, we propose \shortname (\textbf{T}hink \textbf{R}ight with \textbf{A}daptive, \textbf{A}ttentive \textbf{C}ompression), an online post-training RL method that leverages the model's self-attention to identify key steps and prune redundant ones. \shortname also estimates difficulty and incorporates it into training rewards, thereby learning to allocate a reasoning budget commensurate with example difficulty. Across a variety of tasks (\aime, \amc, \gpqa, \bbeh), 
\shortname (\qwen)  achieves an average absolute accuracy gain of 8.4\% with a relative reduction in reasoning length of 36.8\% compared to the base model, and a 7.9\% accuracy gain paired with a 29.4\% length drop compared to the best RL baseline. \shortname generalizes well, with accuracy and efficiency gains on out-of-distribution non-math datasets like \gpqa, \bbeh, and \optimalThinkBench. Our analysis shows that \shortname{} learns to adjust its thinking budget based on difficulty and that a combination of task-difficulty calibration and attention-based compression yields gains across diverse tasks.\footnote{Codebase: \href{https://github.com/joykirat18/TRAAC}{https://github.com/joykirat18/TRAAC}}

\end{abstract}

\section{Introduction}

Recent advancements in thinking models have enabled language models to solve complex reasoning tasks~\citep{deepseekai2025deepseekr1incentivizingreasoningcapability, openai2024openaio1card, qwen3technicalreport}. These models extend the chain-of-thought~\citep[CoT;][]{wei2023chainofthoughtpromptingelicitsreasoning} paradigm with online reinforcement learning~\citep[RL;][]{ shao2024deepseekmathpushinglimitsmathematical}, allowing them to refine intermediate solutions as well as sequentially scaling the number of tokens (i.e., compute) to arrive at the final answer.
While such approaches show strong promise for harder problems in domains like mathematics, programming, and logical puzzles~\citep{xie2025logicrlunleashingllmreasoning, chen2025enigmatascalinglogicalreasoning}, their accuracy and utility remain capped by a failure to regulate their reasoning length. 
On one hand, \textit{underthinking} arises when models terminate too early on harder problems, yielding an incorrect final answer. 
On the other hand, \textit{overthinking} occurs when models think excessively for simpler tasks, inflating test-time computation~\citep{marjanović2025deepseekr1thoughtologyletsthink, wu2025more, cuadron2025danger}, and reducing efficiency. This highlights the need for adaptive thinking~\citep{saha2025system1xlearningbalancefast, chen2024not, snell2024scalingllmtesttimecompute, aggarwal2025l1controllinglongreasoning}, where models dynamically allocate thinking based on difficulty.

\begin{figure}
    \centering
    \includegraphics[width=0.8\linewidth]{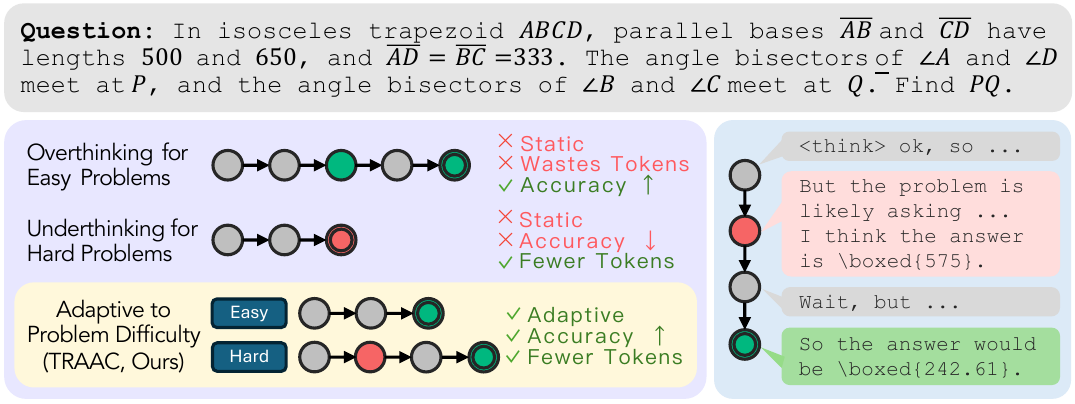}
    \caption{
    Overthinking on easy problems wastes tokens despite being able to maintain decent accuracy. On the other hand, underthinking on hard problems saves token budgets but fails to maintain accuracy. \shortname addresses this trade-off by adapting to problem difficulty (estimated during training), via attention-based compression and, enabling intelligent resource allocation while improving both accuracy and efficiency.}
    \label{fig:fig1}
\end{figure}

We refer to the phenomenon of models misallocating thinking budget -- illustrated in \cref{fig:fig1} -- as \textit{\textbf{under-adaptivity}}. Addressing under-adaptivity is crucial for improving both performance and efficiency of long-thinking models, as dynamic reasoning effort allocation can enable better reasoning exploration in harder problems, while avoiding wasteful computation on problems requiring minimal reasoning. 
Prior work has generally addressed the upper end of under-adaptivity, i.e., improving thinking efficiency.
These works employ supervised fine-tuning on compressed CoT~\citep{xia2025tokenskipcontrollablechainofthoughtcompression}, using user control signals such as early stopping during inference~\citep{muennighoff2025s1simpletesttimescaling}, or RL methods with length penalties~\citep{arora2025traininglanguagemodelsreason, hou2025thinkprune}.
Other more adaptive work has employed budget-aware reward shaping with a binary choice between thinking or not thinking~\citep{zhang2025adaptthinkreasoningmodelslearn}. 
While this can reduce token usage, 
its performance is typically bounded by that of the underlying model, and often it trades performance for efficiency. 
We aim to beat this trade-off and improve both efficiency and accuracy by providing finer-grained feedback via difficulty-adaptive compression, where the compression level is dynamically adapted to task difficulty to address under-adaptivity.

To address these gaps, we introduce \shortname(\textbf{T}hink \textbf{R}ight with \textbf{A}daptive, \textbf{A}ttentive \textbf{C}ompression), a GRPO-based~\citep{shao2024deepseekmathpushinglimitsmathematical} post-training method that incorporates an \textbf{online, difficulty-adaptive, attention-based compression module} to adaptively prune the reasoning trajectory (an entire chain in \cref{fig:fig1}) based on estimated task difficulty. Here online also refers to the fact that task difficulty is re-estimated and the compression rate is recalibrated at every training step from the currently policy's rollout distribution.
Our method teaches the model to compress the context that it should pay attention to, such that it contains only relevant material without getting distracted or skewed in wrong directions~\citep{weston20232attentionisneed}. 
Specifically, we compute the attention score averaged across layers and heads of the model for each reasoning step (\cref{fig:fig1} (right)) from the \texttt{</think>} token and \textit{compress} reasoning steps that are \textit{least attended to}, based on the assumption that these are the least important tokens contributing to the final answer. 
During online training, the level of attention-compression is determined by task difficulty, as estimated by the pass rate during GRPO rollout, making the model more adaptive.  
For harder problems, \shortname maintains a low compression rate, allowing extended reasoning, which increases the likelihood of reaching the correct final answer. For easier problems, it applies a higher compression rate to aggressively compress once the correct final answer is reached.

We evaluate \shortname on two strong off-the-shelf reasoning models, \qwen~\citep{qwen3technicalreport} and \deepseek~\citep{deepseekai2025deepseekr1incentivizingreasoningcapability}, across multiple benchmarks: \amc~\citep{amc23}, \aime~\citep{aime24}, GPQA-Diamond~\citep{rein2023gpqagraduatelevelgoogleproofqa}, \bbeh~\citep[Big Bench Extra Hard;][]{kazemi2025bigbenchextrahard}, and \optimalThinkBench~\citep{aggarwal2025optimalthinkingbenchevaluatingunderthinkingllms}. 
Our experiments demonstrate that \shortname consistently adapts to problem difficulty, yielding improvements in efficiency on simple tasks and stronger accuracy on complex tasks. 
Averaged across \amc, \aime, \gpqa, and \bbeh, \shortname (\qwen) achieves an average absolute improvement of 8.4\% in accuracy while a relative reduction in reasoning length by 36.8\% compared to the base model. Among OOD tasks (\gpqa, \bbeh, \overthinkBench, and \underthinkBench), \shortname shows an average improvement of 3\% on Qwen3-4B, with a maximum improvement of 6.8\% on UnderthinkBench, along with an average 40\% reduction in response length across OOD tasks.  
Our analysis and ablations demonstrate that through difficulty level calibration, \shortname learns to dynamically adjust its compression ratio -- with lower compression on difficult tasks and higher compression on easier ones, yielding performance gains across diverse difficulty tasks.

\section{\shortname: \underline{T}hink \underline{R}ight with \underline{A}daptive \underline{A}ttentive \underline{C}ompression}

\begin{figure*}[t]
\begin{center}
\includegraphics[width=0.8\textwidth]{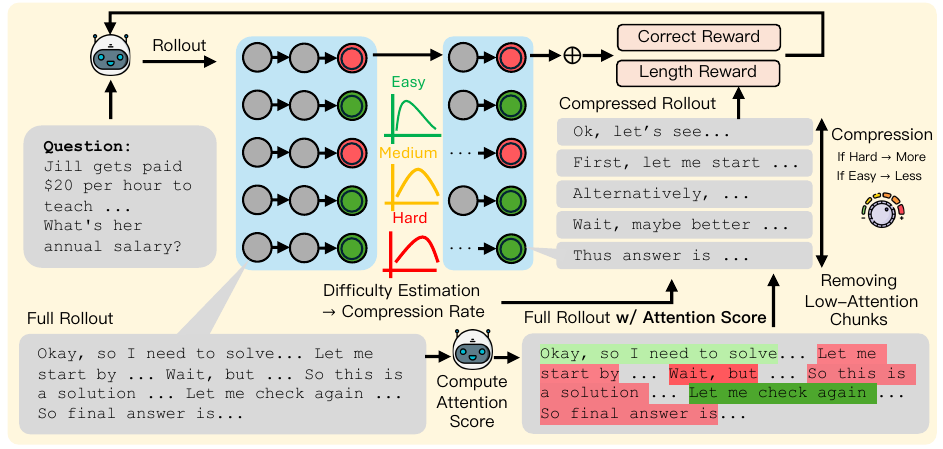}
\end{center}
\caption{Overview of \shortname. Given a problem, the model first generates $N$ rollouts, and the pass rate estimates the problem's difficulty (easy, medium, or hard). Next, the generated reasoning is fed back into the model, which is asked to compute the attention score of each reasoning token from \texttt{</think>}; we then remove steps with lower scores. The degree of removal is determined by the estimated difficulty: easier problems undergo more aggressive compression. Finally, we compute the correctness and length rewards over a group sampled from both the original and compressed rollouts, and use these rewards to update the policy.}
\label{fig:mainFigure}
\end{figure*}

In this section, we introduce our proposed \shortname method in detail (\cref{fig:mainFigure}). It is designed to mitigate under-adaptivity, which leads to resource misallocation during test-time. The main challenge lies in the efficient identification of low-importance tokens and making the attention-based compression adaptive to the task's difficulty. \shortname employs an attention-based compression module that calibrates degree of compression based on estimated task difficulty and prunes unnecessary reasoning steps while preserving essential information.

\subsection{Problem Formulation in \shortname}
\shortname is based on Group Relative Policy Optimization~\citep[GRPO;][]{shao2024deepseekmathpushinglimitsmathematical}, which is an online reinforcement learning (RL) framework that extends Proximal Policy Optimization~\citep{schulman2017proximalpolicyoptimizationalgorithms} by eliminating the critic and instead estimating the baseline from a group of sampled responses. 
Let $\pi_{\theta}$ denote the policy model and $q$ the input query. Given $q$, the model generates an output \(y = \mathrm{cat}(r, a)\) where $\mathrm{cat}$ is the concatenate function, $r$ is the complete reasoning trajectory, and $a$ is the final answer, separated by the delimiter \texttt{</think>}. An attention-based compression module $\mathcal{C}$ (described below) produces a compressed reasoning trajectory:  
$r_{\text{comp}} = \mathcal{C}(r)$. At each training step, the model generates $N$ rollouts, $\{y^i\}_{i=1}^N$, where each rollout $y^i = \mathrm{cat}(r^i, a^i)$ (see ``rollout'' arrow in \cref{fig:mainFigure}). The advantage of each rollout is estimated using the standard GRPO objective (details in Appendix~\ref{appendix:grpo}). The task difficulty $\texttt{d}$ 
is estimated from these rollouts as the proportion of correct answers among the $N$ samples~\citep{zhang2025grpoleaddifficultyawarereinforcementlearning, huang2025adactrladaptivecontrollablereasoning}. 
We show this in \cref{fig:mainFigure} by classifying a problem to easy, medium or hard based on $\texttt{d}$. Task difficulty
\texttt{d} is then used to (i) modulate the compression ratio applied to the reasoning trajectory $r$, and (ii) assign rewards to each rollout. The answer is regenerated based on the compressed trajectory using the
same policy. We then form the GRPO group by uniformly sampling $N$ rollouts from the combined pool of $2N$ original and compressed rollouts, and compute the three rewards on this group. Including both original and compressed rollouts give GRPO a within-group length vs. correctness contrast. When a compressed rollout still reaches the correct answer, its shorter length earns a higher relative advantage than the longer original and pushes the policy toward compression, and when compression breaks correctness, the original is favored instead.
\subsection{Adaptive, Attentive Compression Module}
The goal of the compression module is to identify and remove redundant reasoning steps by evaluating attention scores assigned to each token.

\textbf{Attention-Based Compression.} To calculate the attention score assigned to each token, we pass the reasoning trajectory $r$ (full rollout in \cref{fig:mainFigure}) through the initial policy model. As compared to other compression-based methods~\citep{cheng2025optimizinglengthcompressionlarge, lu2025retro}, \shortname does not rely on external models for annotating reasoning steps. 
To segment the reasoning trajectory $r$ into reasoning steps, we split it at special control tokens such as \textit{``wait''}, \textit{``alternative''}, etc  (complete list Appendix~\ref{app:special_tokens}). 
For the current thinking models, \texttt{</think>} marks the end of a reasoning trajectory, followed by the final answer.~\citet{choi2025thinkclearlyimprovingreasoning} show that \texttt{</think>} attends to key reasoning steps that contain crucial information for deriving the final answer, therefore, for each token $t_j$ in the reasoning steps, its importance score is defined as the aggregated attention from the delimiter \texttt{</think>} across all layers and heads:  
\[
s_j = \frac{1}{LH} \sum_{\ell=1}^L \sum_{h=1}^H 
\alpha_{\texttt{</think>} \to t_j}^{(\ell, h)},
\] 
where $L$ is the number of layers, $H$ is the number of heads per layer, and $\alpha_{\texttt{</think>} \to t_j}^{(\ell, h)}$ is the attention weight from \texttt{</think>} to token $t_j$ in head $h$ of layer $\ell$. 
Before computing the attention score of each token, consistent with prior work~\citep{muennighoff2025s1simpletesttimescaling, choi2025thinkclearlyimprovingreasoning}, we also append an auxiliary prompt at the end of the reasoning trajectory (see Appendix~\ref{appendix:compression_prompt}).
This encourages the model to distill the reasoning process into its most salient steps, enabling the delimiter token \texttt{</think>} to attend to the most informative parts of the reasoning trajectory (highlighted in green).  As shown in \cref{fig:mainFigure} (bottom-right), the model assigns low attention scores to reasoning steps that do not contribute to the final correct answer (highlighted in red), effectively pruning unnecessary cyclic self-corrections and verification loops. Finally, the importance score of a reasoning step $C_k$, consisting of tokens $\{t_j\}_{j \in C_k}$, is computed as the mean of its token-level scores:  
$s_{C_k} = \frac{1}{|C_k|} \sum_{j \in C_k} s_j$.
Steps with lower importance scores are pruned, yielding compressed reasoning trajectory $r^i_{\text{comp}}$.

\textbf{Difficulty-Level Calibration.} 
To address \textbf{under-adaptivity}, the pruning strategy is further adapted to task difficulty, i.e., for easier tasks, a larger proportion of reasoning steps are removed, encouraging the model to condense its reasoning more aggressively (see ``compression'' on the right of \cref{fig:mainFigure}). The difficulty of a task is estimated, based on the pass rate of each problem, during rollout. From the estimated difficulty level, each problem set is categorized among three difficulty levels: easy, medium, and hard, with a higher pass rate indicating easier problems and vice versa.  Each category is assigned a compression rate to determine the degree of redundant steps to prune from the reasoning trajectory, with a higher compression for easier problems and a lower compression for hard problems. In addition, to keep these constraints adaptive to the amount of redundancy in the steps, we calculate the \emph{uniformity} of the attention score distribution. When the distribution of $\{s_j\}$ is close to uniform, indicating that no step or token within a step stands out as significantly more important, the compression rate is reduced to avoid removing potentially useful reasoning steps (see Appendix~\ref{appendix:uniformity_score} on calculating the uniformity score). The difficulty estimate \texttt{d} is further incorporated with the reward calculation described below.

\subsection{Rewards}
\label{sec:rewards}
Following standard GRPO practice of having a verifiable reward system~\citep{shao2024deepseekmathpushinglimitsmathematical}, our setup comprises three reward signals to guide the model to generate correct adaptive length responses based on the difficulty of the task:
\begin{itemize}
    \itemsep0em 
    \item \textbf{Correctness Reward (CR):} A high-weight reward is assigned to outputs that produce the correct final answer. A high score over other rewards is used to ensure that correctness remains the primary optimization objective, regardless of the reasoning trajectory length. 
    \item \textbf{Format Reward:} ensures the presence of special delimiter tokens such as \texttt{<think>} and \texttt{</think>}, ensuring that trajectory $r$ and final answer $a$ are easily distinguishable.
    \item \textbf{Length Reward (LR):} To regulate the verbosity of the reasoning process, we define a length-based reward that penalizes unnecessarily long reasoning traces while adapting to task difficulty. Based on our initial experiment, simply favoring shorter rollouts led to a drastic decrease in response length along with model accuracy; therefore we introduce a sigmoid-based smoothing mechanism that provides a soft bonus ($\beta$) for rollouts beyond the median length. This prevents sharp drops in reward for slightly longer reasoning and helps stabilize training. During each training step, rollouts are partitioned into bins according to their calculated difficulty. As mentioned above, we use the pass rate of the rollouts to categorize them into three difficulty bins: easy, medium and hard. For each bin, we maintain a different distribution  $\mathcal{L}_\texttt{d} = \{\ell_1, \ell_2, \dots, \ell_m\}$ for each difficulty category, where each $\ell_i$ denotes the reasoning length of a rollout within that difficulty category \texttt{d}.  Let $\ell$ be the length of the current rollout. The normalized length score is computed as: \( L_{\text{norm}} = (L_{\max} - \ell) / {\max(L_{\max} - L_{\min}, \epsilon)} \),
    where $\epsilon > 0$ prevents division by zero and \(L_{\min} = \min(\mathcal{L}), L_{\max} = \max(\mathcal{L})\). To avoid a sharp cutoff around the median, we add a smooth bonus term: 
    \[
    \beta = 1/\Big({1 + \exp\!\left(\tfrac{\ell - \text{median}(\mathcal{L})}{0.1 \times \text{median}(\mathcal{L})}\right)\Big)},
    \]
    where \(\text{median}(\mathcal{L})\!= \!\text{median of the set}.\) The final length reward becomes $r_{\text{length}}\!= \max\!\left(L_{\text{norm}}, \ \beta\right)$.
    Note that length reward is only provided to a rollout if it reaches a final correct answer. Moreover, to ensure stability when calculating $L_{min}$, $L_{max}$, and medium($\mathcal{L}$), we maintain a sliding window over the last 10 steps for each difficulty bin, thereby avoiding drastic fluctuations during training. 
\end{itemize}
The final reward for each rollout during GRPO training is the combination of correctness, format, and length rewards (c.f. range of each reward in Appendix~\ref{app:training_reward}).

\begin{table*}[t!]

\centering
\small
\begin{tabular}{lcccccccccc}
\toprule
\multicolumn{1}{l|}{\textbf{Method}}           & \multicolumn{2}{c|}{\textbf{\aime}}         & \multicolumn{2}{c|}{\textbf{\amc}}          & \multicolumn{2}{c|}{\textbf{\gpqa}}         & \multicolumn{2}{c|}{\textbf{\bbeh}}         & \multicolumn{2}{c}{\textbf{Average}} \\ \midrule
\multicolumn{1}{l|}{}                          & Acc.$\uparrow$           & \multicolumn{1}{c|}{Len.$\downarrow$} & Acc.$\uparrow$           & \multicolumn{1}{c|}{Len.$\downarrow$} & Acc.$\uparrow$           & \multicolumn{1}{c|}{Len.$\downarrow$} & Acc.$\uparrow$           & \multicolumn{1}{c|}{Len.$\downarrow$} & Acc.$\uparrow$                  & Len.$\downarrow$         \\ \midrule
\multicolumn{11}{c}{\textbf{Qwen3-4B}}                                                                                                                                                                                                                               \\ \midrule
\multicolumn{1}{l|}{Base Model}                & 27.64          & \multicolumn{1}{c|}{9.2}  & 68.19          & \multicolumn{1}{c|}{7.0}  & 45.18          & \multicolumn{1}{c|}{7.6}  & 18.28          & \multicolumn{1}{c|}{6.7}  & 39.8                  & 7.6          \\
\multicolumn{1}{l|}{TokenSkip}                 & 5.84           & \multicolumn{1}{c|}{9.6}  & 27.71          & \multicolumn{1}{c|}{8.7}  & 32.32          & \multicolumn{1}{c|}{7.8}  & 11.91          & \multicolumn{1}{c|}{7.2}  & 19.4                  & 8.3          \\
\multicolumn{1}{l|}{L1-Max}                    & 30.11          & \multicolumn{1}{c|}{7.1}  & 63.61          & \multicolumn{1}{c|}{5.8} & 43.23          & \multicolumn{1}{c|}{5.8}  & 14.91          & \multicolumn{1}{c|}{5.0}  & 38.0                  & 5.9          \\
\multicolumn{1}{l|}{LC-R1}                     & 13.48          & \multicolumn{1}{c|}{2.6}  & 56.38          & \multicolumn{1}{c|}{1.7}  & 26.67          & \multicolumn{1}{c|}{1.5}  & 12.35          & \multicolumn{1}{c|}{1.9}  & 27.2                  & 1.9          \\
\multicolumn{1}{l|}{Adapt Think}               & 36.63          & \multicolumn{1}{c|}{8.4}  & 72.77          & \multicolumn{1}{c|}{5.8}  & 44.04          & \multicolumn{1}{c|}{6.7}  & 7.87           & \multicolumn{1}{c|}{6.2}  & 40.3                  & 6.8          \\
\multicolumn{1}{l|}{\shortname} & \textbf{45.45} & \multicolumn{1}{c|}{6.7}  & \textbf{79.52} & \multicolumn{1}{c|}{4.2}  & \textbf{47.21} & \multicolumn{1}{c|}{4.2}  & \textbf{20.59} & \multicolumn{1}{c|}{4.3}  & \textbf{48.2}         & 4.8          \\ \midrule
\multicolumn{11}{c}{\textbf{DeepSeek-R1-Distill-Qwen-7B}}                                                                                                                                                                                                     \\ \midrule
\multicolumn{1}{l|}{Base Model}                & 33.71          & \multicolumn{1}{c|}{8.2}  & 74.22          & \multicolumn{1}{c|}{5.7}  & 43.55          & \multicolumn{1}{c|}{7.1}  & 10.61          & \multicolumn{1}{c|}{5.9}  & 40.5                  & 6.7          \\
\multicolumn{1}{l|}{TokenSkip}                 & 24.94          & \multicolumn{1}{c|}{8.5}  & 52.05          & \multicolumn{1}{c|}{6.8}  & 34.24          & \multicolumn{1}{c|}{7.0}  & 6.30           & \multicolumn{1}{c|}{6.4}  & 29.4                  & 7.2          \\
\multicolumn{1}{l|}{L1-Max}                    & 31.01          & \multicolumn{1}{c|}{3.1}  & 75.90          & \multicolumn{1}{c|}{2.2}  & 23.54          & \multicolumn{1}{c|}{1.9}  & \textbf{13.43} & \multicolumn{1}{c|}{2.1}  & 36.0                  & 2.3          \\
\multicolumn{1}{l|}{LC-R1}                     & 6.07           & \multicolumn{1}{c|}{4.0}  & 37.35          & \multicolumn{1}{c|}{3.5}  & 28.78          & \multicolumn{1}{c|}{2.5}  & 9.09           & \multicolumn{1}{c|}{1.7}  & 20.3                  & 2.9          \\
\multicolumn{1}{l|}{Adapt Think}               & \textbf{38.88} & \multicolumn{1}{c|}{7.1}  & 75.66          & \multicolumn{1}{c|}{4.1}  & 19.29          & \multicolumn{1}{c|}{4.8}  & 6.17           & \multicolumn{1}{c|}{5.2}  & 35.0                  & 5.3          \\
\multicolumn{1}{l|}{\shortname} & 38.60          & \multicolumn{1}{c|}{7.3}  & \textbf{77.83} & \multicolumn{1}{c|}{4.5}  & \textbf{47.31} & \multicolumn{1}{c|}{6.2}  & 11.55          & \multicolumn{1}{c|}{5.2}  & \textbf{43.8}         & 5.8          \\ \bottomrule
\end{tabular}
\caption{Performance comparison of \shortname with various baselines. Each model is evaulate across various benchmarks, and Acc: accuracy(\%) and Len: average Response Length(k) are reported. \shortname on average shows the highest performance gain. 
} 
\label{tab:mainResults}
\end{table*}
\section{Experimental Setup}
\label{sec:experimentSetup}
\textbf{Models.} We adopt two reasoning models, DeepSeek-R1-Distill-Qwen-7B~\citep{deepseekai2025deepseekr1incentivizingreasoningcapability} (\deepseek) and Qwen3-4B~\citep{qwen3technicalreport} as our base models.

\textbf{Datasets.} We train the model using DAPO-Math-17k~\citep{yu2025dapoopensourcellmreinforcement} and for evaluation, we use a diverse set of benchmarks, including \aime~\citep{aime24}, \amc~\citep{amc23},
\gpqa~\citep{rein2023gpqagraduatelevelgoogleproofqa}, \overthinkBench/ \underthinkBench~\citep{aggarwal2025optimalthinkingbenchevaluatingunderthinkingllms}, and Big Bench Extra Hard (\bbeh)~\citep{kazemi2025bigbenchextrahard}. Only \aime and \amc are math-specific; the remaining benchmarks represent out-of-distribution settings.
Further dataset details and their sizes are provided in Appendix~\ref{appendix:dataset}.

\textbf{Training and Evaluation.} We set temperature to 1.0 and maximum response length to 10k for both training and evaluation. During GRPO rollout, we sample 8 rollouts per step, and report mean accuracy and response length across 5 evaluation runs. For difficulty calibration, we bin problems into easy, medium, and hard categories, with decreasing compression scores (see Appendix~\ref{appendix:hyperparameters} for other hyperparameter details). For evaluating the overthinking split, we also report the $\text{AUC}_{\text{OAA}}$~\citep{aggarwal2025optimalthinkingbenchevaluatingunderthinkingllms}. Intuitively, a higher $\text{AUC}_{\text{OAA}}$ indicates that the model sustains stronger accuracy while minimizing unnecessary reasoning across thresholds. For \optimalThinkBench score, we combined the $\text{AUC}_{\text{OAA}}$ from \overthinkBench and accuracy from \underthinkBench into a single F1 score. More details about these metrics are in Appendix~\ref{appendix:eval_metrics}.

\textbf{Baselines.} We compare \shortname with 5 strong baselines:
\textbf{(1) Base model:} off-the-shelf reasoning model, 
\textbf{(2) TokenSkip:} An SFT based baseline as described by ~\citet{xia2025tokenskipcontrollablechainofthoughtcompression} that fine-tunes the model over compressed CoT training data.
\textbf{(3) L1-Max}: An RL framework proposed by ~\citet{aggarwal2025l1controllinglongreasoning} that optimizes for accuracy while adhering to user-specific length constraints. We used the constraint  ``Think for a maximum of 10000 tokens.'' during its training.
\textbf{(4) LC-R1:} A compression-based RL framework by ~\citet{cheng2025optimizinglengthcompressionlarge} that uses an externally trained model to remove invalid portions of
the thinking process.
\textbf{(5) AdaptThink:} an adaptive RL framework described by ~\citet{zhang2025adaptthinkreasoningmodelslearn}, that enables reasoning models to choose between ``thinking'' and ``no-thinking''  modes and poses it as a constraint optimization problem that encourages the model to choose no-thinking while maintaining performance. Prompts used for all baselines in Appendix~\ref{app:Baseline_prompt}.

\section{Result and Discussion}
\subsection{Main Results}
\label{sec: results}

\textbf{\shortname improve both performance and efficiency.}
Tables~\ref{tab:mainResults} compares TRAAC against baselines on \aime, \amc, \gpqa, and \bbeh. \shortname (\qwen)  achieves an average accuracy improvement of 8.4\% while reducing reasoning length by 36.8\% compared to the base model. Similarly, \shortname (\deepseek) improves accuracy by 3.3\% with a 13.4\% reduction in length. When compared to the SFT baseline TokenSkip~\citep{xia2025tokenskipcontrollablechainofthoughtcompression}, \shortname outperforms in terms of performance and efficiency for both models, \qwen and \deepseek. Appendix~\ref{app:why_rl} further highlights why RL is necessary to teach adaptive compression rather than SFT over compressed rollouts.
Similarly, L1-Max~\citep{aggarwal2025l1controllinglongreasoning}, an RL-based method that penalizes long responses, also solely focuses on efficiency gains, at a slight cost of overall performance. 
Additionally, the compression-based RL framework LC-R1~\citep{cheng2025optimizinglengthcompressionlarge} improves the efficiency of the model at the cost of a 12.6\% drop for \qwen and 20.2\% drop for \deepseek, when compared with base models, respectively. 
On average for \qwen, \shortname outperforms L1-Max by 10.2\% on \qwen and by 7.9\% on \deepseek. Similarly, \shortname also outperforms LC-R1 by 21\% on \qwen and 23\% on \deepseek. Moreover, given the same token budget, of approximately 7k, \shortname (\qwen) on \aime outperforms L1-Max by 15\%. These results highlight that, unlike methods that prioritize only efficiency, \shortname simultaneously delivers both higher accuracy and shorter reasoning traces. 

\textbf{\shortname generalizes across domains.} 
\shortname trains on data from DAPO-Math-17k~\citep{yu2025dapoopensourcellmreinforcement}, which is a math reasoning dataset. 
We also evaluate \shortname on several out-of-domain (OOD) tasks, including \gpqa, \bbeh, \overthinkBench, and \underthinkBench (Table~\ref{tab:optimal_main}). Among these OOD tasks, \shortname shows an average improvement of 3\% on \qwen and 2.8\% on \deepseek compared to the base model, with improvement as high as 6.8\% on \underthinkBench, which covers 100 diverse reasoning tasks from Reasoning Gym~\citep{stojanovski2025reasoninggymreasoningenvironments}. In addition, \shortname reduces reasoning tokens by 40\% on \qwen and 20\% on \deepseek, demonstrating substantially higher efficiency while also boosting accuracy across benchmarks. In Appendix~\ref{app:agentic_Eval}, we also show \shortname's improved performance and efficiency on agentic, multi-turn benchmark that requires multi-step interactions and tool use. This indicates that \shortname learns a generalizable compression strategy that transfers from math to other reasoning domains.

\begin{table}[!t]

\centering
{\small
\begin{tabular}{lcccccc}
\toprule
\multicolumn{1}{l|}{\textbf{Method}}           & \multicolumn{3}{c|}{\textbf{\overthinkBench}}                     & \multicolumn{2}{c|}{\textbf{\underthinkBench}}   & \textbf{\optimalThinkBench} \\ \midrule
\multicolumn{1}{l|}{}                          & Acc.$\uparrow$           & Len.$\downarrow$ & \multicolumn{1}{c|}{$\text{AUC}\uparrow$}            & Acc.$\uparrow$           & \multicolumn{1}{c|}{Len.$\downarrow$} & F1$\uparrow$                            \\ \midrule
\multicolumn{7}{c}{\textbf{Qwen3-4B}}                                                                                                                                                \\ \midrule
\multicolumn{1}{l|}{Base Model}                & \textbf{90.02} & 1.2  & \multicolumn{1}{c|}{80.06}          & 34.33          & \multicolumn{1}{c|}{7.1}  & 48.05                         \\
\multicolumn{1}{l|}{TokenSkip}                 & 78.15          & 3.5  & \multicolumn{1}{c|}{57.88}          & 14.80          & \multicolumn{1}{c|}{7.9}  & 23.57                         \\
\multicolumn{1}{l|}{L1-Max}                    & 87.22          & 0.9  & \multicolumn{1}{c|}{1.11}           & 21.27          & \multicolumn{1}{c|}{6.3}  & 2.10                          \\
\multicolumn{1}{l|}{LC-R1}                     & 78.62          & 0.3  & \multicolumn{1}{c|}{64.20}          & 14.95          & \multicolumn{1}{c|}{1.3}  & 24.25                         \\
\multicolumn{1}{l|}{Adapt Think}               & 68.83          & 8.2  & \multicolumn{1}{c|}{63.44}          & 18.80          & \multicolumn{1}{c|}{6.0}  & 29.01                         \\
\multicolumn{1}{l|}{\shortname} & 89.79          & 0.6  & \multicolumn{1}{c|}{\textbf{85.06}} & \textbf{41.09} & \multicolumn{1}{c|}{4.7}  & \textbf{55.41}                \\ \midrule
\multicolumn{7}{c}{\textbf{DeepSeek-R1-Distill-Qwen-7B}}                                                                                                                      \\ \midrule
\multicolumn{1}{l|}{Base Model}                & 78.45          & 0.9  & \multicolumn{1}{c|}{72.38}          & 12.69          & \multicolumn{1}{c|}{6.2}  & 21.60                         \\
\multicolumn{1}{l|}{TokenSkip}                 & 57.03          & 3.9  & \multicolumn{1}{c|}{40.77}          & 8.55           & \multicolumn{1}{c|}{7.2}  & 14.13                         \\
\multicolumn{1}{l|}{L1-Max}                    & 73.18          & 1.0  & \multicolumn{1}{c|}{66.01}          & 20.07          & \multicolumn{1}{c|}{2.0}  & 30.78                         \\
\multicolumn{1}{l|}{LC-R1}                     & 76.08          & 0.9  & \multicolumn{1}{c|}{69.81}          & 7.16           & \multicolumn{1}{c|}{2.5}  & 12.99                         \\
\multicolumn{1}{l|}{Adapt Think}               & 73.41          & 0.4  & \multicolumn{1}{c|}{70.72}          & 13.13          & \multicolumn{1}{c|}{4.6}  & 22.14                         \\
\multicolumn{1}{l|}{\shortname} & \textbf{81.81} & 1.0  & \multicolumn{1}{c|}{\textbf{72.89}} & \textbf{22.30} & \multicolumn{1}{c|}{5.9}  & \textbf{34.15}                \\ \bottomrule
\end{tabular}
}
\caption{Performance of \shortname and various baselines on \optimalThinkBench. For \underthinkBench we report the Acc: Accuracy(\%), and Len: Average Response length(k). For \overthinkBench, in addition to Acc. and Len. we also report the $\text{AUC}_{\text{OAA}}$.}
\label{tab:optimal_main}
\vspace{-10pt}
\end{table}

\textbf{\shortname learns to adaptively allocate token budget.} 
Among the baselines in Tables~\ref{tab:mainResults} and ~\ref{tab:optimal_main}, we also compare \shortname against an adaptive RL method, AdaptThink~\citep{zhang2025adaptthinkreasoningmodelslearn}, which teaches the model to use distinct ``thinking'' vs. ``non-thinking'' modes for hard and easy problems, respectively. On \qwen, \shortname outperforms AdaptThink by 7.9\% while also reducing tokens by 29.4\%, highlighting that a flexible adaptive strategy is more effective in handling diverse problem difficulties. 
Table~\ref{tab:optimal_main} further tests on the \overthinkBench/\underthinkBench~\citep{aggarwal2025optimalthinkingbenchevaluatingunderthinkingllms}. \overthinkBench measures excessive use of thinking tokens on simple queries. While, \underthinkBench evaluates how necessary ``thinking'' is based on problem difficulty. 
Taken together, \shortname improves overall F1 performance by 7.36\% on \qwen, and 12.55\% on \deepseek over base model, indicating that \shortname enables the model to avoid both overthinking on simple problems and underthinking on complex ones~\citep{aggarwal2025optimalthinkingbenchevaluatingunderthinkingllms}.
Against AdaptThink, \shortname achieves a 26\% gain on \qwen and a 12\% gain on \deepseek, underscoring its ability to adaptively allocate reasoning effort and adjust token budgets based on problem difficulty.
On \overthinkBench, we measure overthinking using the $\text{AUC}{\text{OAA}}$ metric, which rewards models that solve very easy problems correctly while using minimal tokens (ideally 0). Compared to the base model, \shortname (\qwen) improves $\text{AUC}{\text{OAA}}$ by 5\% and \deepseek by 0.5\%. Relative to AdaptThink, \shortname gains 21.6\% for \qwen and 6.9\% for \deepseek.

\subsection{Ablations and Analysis}
\label{sec:analysis}
\begin{figure*}
    \centering
    \includegraphics[width=0.95\textwidth]{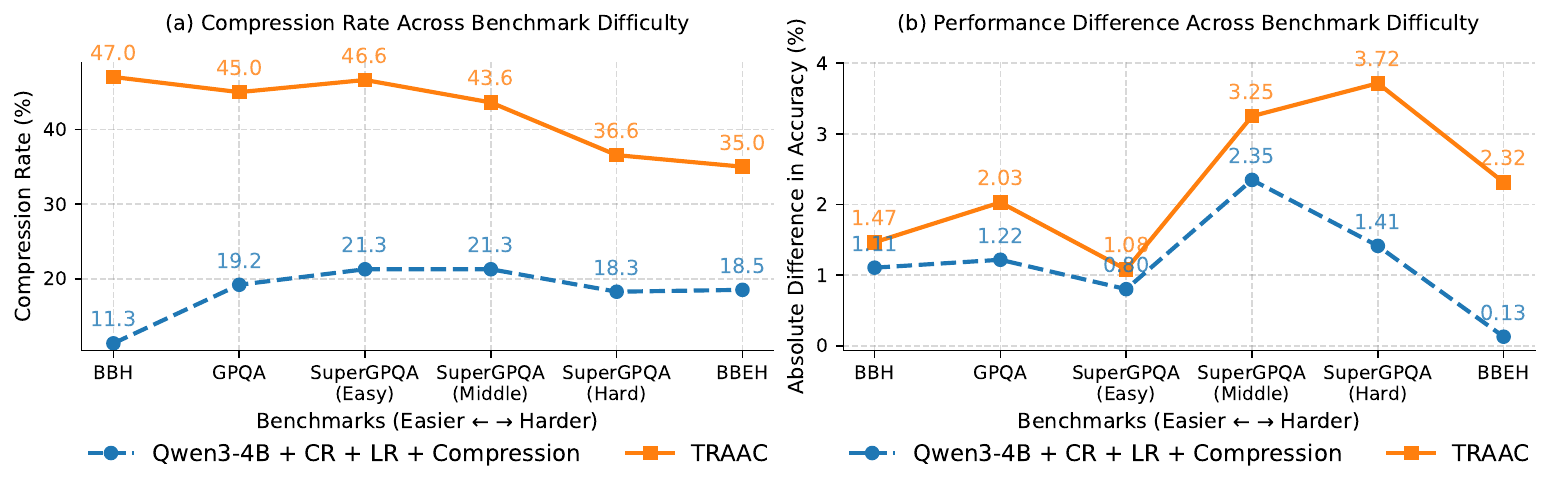}
    \caption{(a) Relative change in compression rate of \shortname and \qwen + Compression compared to \qwen across varying problem difficulty. (b) Absolute accuracy drop of \shortname and \qwen + Compression compared to \qwen across varying difficulty.
    }
    \label{fig:analysis}
\end{figure*}
To understand the importance of each component, we conducted an ablation study, removing each component of our method. 
\cref{tab:ablation} and ~\cref{tab:optimal_ablation} show the performance of these ablations compared with the base model. Specifically, we start with the base model and the ablations: \textbf{(i) Base Model + CR:} The base model trained with GRPO using only the correctness reward, \textbf{(ii) Base model + CR + LR:} The base model trained with GRPO using both correctness and length rewards, but without difficulty-level calibration, \textbf{(iii) Base model + CR + LR + Compression:} The base model trained with GRPO using correctness and length rewards, along with the compression module, with no difficulty-level calibration. We also ablate the compression signal, finding that attention-based pruning outperforms both random and confidence-based step pruning(see Appendix~\ref{app:attention}).
\begin{table*}
\centering

{\small
\resizebox{1.0\textwidth}{!}{%
\begin{tabular}{lcccccccccc}
\toprule
\multicolumn{1}{l|}{\textbf{Method}}           & \multicolumn{2}{c|}{\textbf{\aime}}         & \multicolumn{2}{c|}{\textbf{\amc}}          & \multicolumn{2}{c|}{\textbf{\gpqa}}         & \multicolumn{2}{c|}{\textbf{\bbeh}}         & \multicolumn{2}{c}{\textbf{Avg.}} \\ \midrule
\multicolumn{1}{l|}{}                          & Acc.$\uparrow$           & \multicolumn{1}{c|}{Len.$\downarrow$} & Acc.$\uparrow$           & \multicolumn{1}{c|}{Len.$\downarrow$} & Acc.$\uparrow$           & \multicolumn{1}{c|}{Len.$\downarrow$} & Acc.$\uparrow$           & \multicolumn{1}{c|}{Len.$\downarrow$} & Acc.$\uparrow$                  & Len.$\downarrow$         \\ \midrule
\multicolumn{11}{c}{\textbf{Qwen3-4B}}                                                                                                                                                                                                                               \\ \midrule
\multicolumn{1}{l|}{Base Model}                & 27.64          & \multicolumn{1}{c|}{9.2}  & 68.19          & \multicolumn{1}{c|}{7.0}  & 45.18          & \multicolumn{1}{c|}{7.6}  & 18.28          & \multicolumn{1}{c|}{6.7}  & 39.8                  & 7.6          \\
\multicolumn{1}{l|}{\: + CR}           & 44.36          & \multicolumn{1}{c|}{7.9}  & 77.35          & \multicolumn{1}{c|}{5.5}  & 46.29          & \multicolumn{1}{c|}{5.7}  & 18.13          & \multicolumn{1}{c|}{5.2}  & 46.5                  & 6.1          \\
\multicolumn{1}{l|}{\: + LR}      & 37.84          & \multicolumn{1}{c|}{4.5}  & 77.35          & \multicolumn{1}{c|}{2.4}  & 44.06          & \multicolumn{1}{c|}{2.3}  & 18.57          & \multicolumn{1}{c|}{2.1}  & 44.5                  & 2.8          \\
\multicolumn{1}{l|}{\: + Compression}  & 38.37          & \multicolumn{1}{c|}{8.1}  & 75.90          & \multicolumn{1}{c|}{5.5}  & 46.40          & \multicolumn{1}{c|}{6.2}  & 18.41          & \multicolumn{1}{c|}{5.4}  & 44.8                  & 6.3          \\ 
\multicolumn{1}{l|}{\shortname} & \textbf{45.45} & \multicolumn{1}{c|}{6.7}  & \textbf{79.52} & \multicolumn{1}{c|}{4.2}  & \textbf{47.21} & \multicolumn{1}{c|}{4.2}  & \textbf{20.59} & \multicolumn{1}{c|}{4.3}  & \textbf{48.2}         & 4.8          \\ \bottomrule
\end{tabular}
}}
\caption{Ablation Results of \shortname(\qwen, \deepseek) tested across 4 datasets: \aime, \amc, \gpqa, and \bbeh. Each component addition adds to the previous method. }
\label{tab:ablation}
\end{table*}

\textbf{Combining difficulty-adaptiveness and attention-based compression is crucial for accuracy and efficiency.} 
\cref{tab:ablation} shows that removing the difficulty-based calibration (Base Model + CR + LR + compression) reduces the average performance across by 3.4\%, while also making the model less efficient by 23.8\%. 
Additionally, removing the attention-based compression (Base Model + LR + CR) drops performance by an additional 0.3\%. On \optimalThinkBench(\cref{tab:optimal_ablation}): the F1 score decreases by 5.2\% when task-difficulty level calibration is removed and drops further by 7.6\% when the attention-based compression module is also removed. These results confirm that both components are essential for achieving accuracy and efficiency gains across tasks.

\begin{table}
\small
\centering

{\small
\begin{tabular}{lcccccc}
\toprule
\multicolumn{1}{l|}{\textbf{Method}}           & \multicolumn{3}{c|}{\textbf{\overthinkBench}}                     & \multicolumn{2}{c|}{\textbf{\underthinkBench}}   & \textbf{\optimalThinkBench} \\ \midrule
\multicolumn{1}{l|}{}                          & Acc.$\uparrow$           & Len.$\downarrow$ & \multicolumn{1}{c|}{$\text{AUC}\uparrow$}            & Acc.$\uparrow$           & \multicolumn{1}{c|}{Len.$\downarrow$} & F1$\uparrow$                            \\ \midrule
\multicolumn{7}{c}{\textbf{Qwen3-4B}}                                                                                                                                                \\ \midrule
\multicolumn{1}{l|}{Base Model}                & 90.02          & 1.2  & \multicolumn{1}{c|}{80.06}          & 34.33          & \multicolumn{1}{c|}{7.1}  & 48.1                          \\
\multicolumn{1}{l|}{\: + CR}           & 90.02          & 0.9  & \multicolumn{1}{c|}{78.86}          & 37.06          & \multicolumn{1}{c|}{5.7}  & 50.4                          \\
\multicolumn{1}{l|}{\: + LR}      & \textbf{90.94} & 0.4  & \multicolumn{1}{c|}{75.86}          & 29.62          & \multicolumn{1}{c|}{2.3}  & 42.6                          \\
\multicolumn{1}{l|}{\: + Compression}  & 90.12          & 0.9  & \multicolumn{1}{c|}{80.41}          & 36.51          & \multicolumn{1}{c|}{6.0}  & 50.2                          \\
\multicolumn{1}{l|}{\shortname} & 89.79          & 0.6  & \multicolumn{1}{c|}{\textbf{85.06}} & \textbf{41.09} & \multicolumn{1}{c|}{4.7}  & \textbf{55.4}                 \\ \bottomrule
\end{tabular}
}
\caption{Ablation Results of \shortname (\qwen and \deepseek) on \optimalThinkBench. Each component addition adds to the previous method.}
\label{tab:optimal_ablation}
\end{table}

\textbf{\shortname{} adapts to task difficulty.} To further understand the level of adaptivity of \shortname compared to other methods, we plot the relative compression ratio and absolute accuracy gains (w.r.t. the base model) in \cref{fig:analysis} as a function of task difficulty, ranking tasks in order of increasing difficulty.
We conduct these experiments on SuperGPQA~\citep{pteam2025supergpqascalingllmevaluation}, stratified into easy, medium, and hard splits, and BBH (Big Bench Hard)~\citep{suzgun2022challengingbigbenchtaskschainofthought} -- an easier version of BBEH.
To get oracle difficulty ratings, we rank the datasets by the performance of frontier models on them~\citep{kazemi2025bigbenchextrahard, pteam2025supergpqascalingllmevaluation}, with harder datasets being those with lower performance.
From \cref{fig:analysis}(a), as difficulty increases, from left to right, the compression rate steadily drops for \shortname, underscoring its ability to compress more for easier tasks and less for difficult tasks. However, without task-difficulty level calibration, the compression rate remains roughly uniform across the tasks. \cref{fig:analysis}(b) shows that even with more compression, \shortname always maintains higher accuracy than \qwen + CR + LR + compression, confirming the effectiveness of difficulty-adaptive thinking.
Moreover, most of the accuracy gains stems from harder problems, indicating the average accuracy gains seen in \cref{tab:mainResults} come from difficulty-adaptive thinking.  \deepseek results (Appendix~\ref{appendix:deepseek_analysis}) follow a similar trend as \qwen.

\textbf{\shortname's boost performance with minimal computation overhead.} We analyse \shortname's computation cost compared to existing RL baselines. \shortname introduces an attention-based compression in addition to standard generation, and as a result, \shortname utilizes more training time in the initial training steps (Appendix~\ref{app:training_time}). However, the additional training time quickly amortizes as the training progresses (Appendix~\ref{app:progress_cost}). As the model learns to generate shorter, training time steadily decreases and consistently becomes more efficient than baseline in later stages of training. In terms of FLOPs, \shortname reduces inference FLOPs by 27.3\% compared to a GRPO baseline with correctness reward only when evaluated on \amc dataset. These results highlights that while \shortname introduces a small upfront computational overhead during early training, it amortizes rapidly and leads to lower overall computation, particularly at inference time (Full experimental details in  Appendix~\ref{app:cost}).

\section{Related Work}

In the past years, reasoning performance of language models has vastly improved via the introduction of chain-of-thoughts~\citep{wei2023chainofthoughtpromptingelicitsreasoning}, parallel scaling through self-consistency~\citep{wang2023selfconsistency}, and best-of-$N$ sampling~\citep{lightman2023let}. More recently, several works have found sequential scaling -- i.e., increasing the number of reasoning tokens -- to be the most effective approach~\citep{muennighoff2025s1simpletesttimescaling}, especially when combined with online reinforcement learning or distillation from such models~\citep{aggarwal2025l1controllinglongreasoning, shao2024deepseekmathpushinglimitsmathematical, deepseekai2025deepseekr1incentivizingreasoningcapability}. Consequently, the area of efficient reasoning -- maintaining high performance from sequential scaling with minimal token usage -- has become a central research focus~\citep{chen2024not,marjanović2025deepseekr1thoughtologyletsthink, wu2025more}. To this end, prior works compress or prune chain-of-thoughts via early exiting~\citep{zhang2025reasoning,fu2025deep}, train models under pre-specified budgets~\citep{aggarwal2025l1controllinglongreasoning}, learn thoughts latently without generating them~\citep{hao2025training}, use supervised finetuning to avoid overthinking~\citep{xia2025tokenskipcontrollablechainofthoughtcompression, cheng2025optimizinglengthcompressionlarge, lu2025retro}, or add length-based penalties for conciseness~\citep{arora2025traininglanguagemodelsreason, hou2025thinkprune}. However, this line of work does not \emph{explicitly} account for varying problem difficulty, instead relying on the model to learn to allocate budget implicitly; in contrast, \shortname{} introduces difficulty-based supervision for budget allocation. Moreover, prior approaches typically address only overthinking -- reducing output length at the cost of performance drops -- whereas we tackle both over- and underthinking.

Improving \emph{both} reasoning performance and efficiency requires a more \emph{adaptive} approach through explicit training. Prior work such as \citet{zhang2025adaptthinkreasoningmodelslearn} frames adaptivity as a binary decision of \emph{whether} to think, whereas we argue that for harder problems it must involve deciding \emph{how much} to think -- and empirically outperform this baseline in \cref{sec: results}. A similar insight appears in planning, where \citet{saha2025system1xlearningbalancefast} show that mixing ``System 1'' and ``System 2'' reasoning within the same instance outperforms a binary choice between them. \citet{shen2025dastdifficultyadaptiveslowthinkinglarge} pursue difficulty-adaptive training via repeated sampling and offline preference optimization to prefer shorter responses. In contrast, \shortname{} provides attention-based supervision in the compression module through online RL~\citep{deepseekai2025deepseekr1incentivizingreasoningcapability}. Unlike concurrent work by \citet{choi2025thinkclearlyimprovingreasoning}, who prune redundant tokens post hoc, our method adapts compression during training itself -- yielding difficulty-aware reasoning and improved test-time efficiency without generating unnecessary tokens.

\vspace{-5pt}
\section{Conclusion}
\vspace{-5pt}
We introduced \shortname, a post-training RL method that operates online and uses a difficulty-adaptive, attention-based compression module. Through its adaptive attentive compression, \shortname is able to prune its reasoning steps adaptively based on the task difficulty. \shortname addresses the issue of under-adaptivity, which helps improve both performance and efficiency, as thinking longer on harder problems helps in better exploration, and thinking shorter on easier problems avoids wasting of test-time compute. 
Moreover, our method also shows strong generalizability, with evaluation done on various OOD tasks. Through our analysis and ablation, we further verify that our adaptive method can provide fine-grained adjustments to the thinking budget based on the difficulty of the problem, and a combination of task-difficulty calibration and attention-based compression helped achieve both accuracy and efficiency gains.

\bibliography{colm2026_conference}
\bibliographystyle{colm2026_conference}

\appendix
\crefalias{section}{appendix}
\section{Dataset Details}
\label{appendix:dataset}
We evaluated the model on various benchmarks:
\begin{itemize}[topsep=1pt, noitemsep,leftmargin=*]
    \item \amc: All questions come from AMC12 2022, AMC12 2023, and have been extracted from the AOPS wiki page. Total Count: 83
    \item \aime: All questions come from AIME 22, AIME 23, and AIME 24, and have been extracted directly from the AOPS wiki page. Total Count: 90
    \item \gpqa: It is a multiple-choice dataset covering physics, biology, and chemistry. Total Count: 198
    \item \bbeh:  A benchmark designed to push the boundaries of LLM reasoning evaluation. BBEH replaces each task in BBH with a novel task that probes a similar reasoning capability but exhibits significantly increased difficulty. Total Count: 460
    \item \optimalThinkBench: A unified benchmark that jointly evaluates overthinking and underthinking in LLMs and also encourages the development of optimally-thinking models that balance performance and efficiency. Two sub benchmarks: \overthinkBench, featuring simple queries in 72 domains, and \underthinkBench, containing 11 challenging reasoning tasks. \underthinkBench count: 550, \overthinkBench count: 607. 
    \item \bbh: a suite of 23 challenging BIG-Bench tasks. Total Count: 2115
    \item SuperGPQA: A comprehensive benchmark designed to evaluate the knowledge and reasoning abilities of Large Language Models (LLMs) across 285 graduate-level disciplines. Each problem is also categorized as easy, medium and hard. 540 problems for each difficulty category, so the total count is 1620.
\end{itemize}

To calculate the accuracy, we adopt Math-Verify.\footnote{\href{https://github.com/huggingface/Math-Verify}{Huggingface Math-Verify}} For \underthinkBench accuracy calculation, we used the evaluation scripts from Reasoning-Gym~\citep{stojanovski2025reasoninggymreasoningenvironments}

\section{Computational cost analysis of training \shortname}
\label{app:cost}
To understand the computational cost of training \shortname vs other RL-based methods (L1-max, AdaptThink), we compare \shortname with RL baselines on training time and FLOPs.

\subsection{Training Time}
\label{app:training_time}
The GRPO algorithm mainly consists of three stages: (i) \textbf{Rollout}: the LLM produces multiple responses for a given prompt; (ii) \textbf{Scoring}: a scalar reward is assigned to each response; (iii) \textbf{Policy optimisation}: the LLM is updated by optimising the total objective. Since we use the \texttt{math-verify} library -- a rule-based expression system that does not require additional LLM calls for reward computation -- the cost of scoring is negligible.

Table~\ref{tab:grpo_time} reports the breakdown of training time for \shortname compared to RL baselines. For each method, we show the wall-clock time (in seconds) for the first training step, split into rollout time, policy optimisation time, and total time.
\begin{itemize}[topsep=1pt, noitemsep,leftmargin=*]
    \item \textbf{Base Model + CR}: GRPO with correctness reward only.  
    \item \textbf{Base Model + CR + LR}: GRPO with correctness and length rewards (no difficulty calibration).  
\end{itemize}

Other RL baselines such as \textsc{L1-Max} and \textsc{AdaptThink} are also variants of Base Model + CR with an additional length reward term.

\begin{table}[!h]
\small

\centering
\begin{tabular}{lcccc}
\toprule
\textbf{Method} & \textbf{Rollout (sec)} & \textbf{Optimise Policy (sec)} & \textbf{Total Time (sec)} & \textbf{Hardware} \\
\midrule
Base Model + CR         & 250 & 87.5 & 397.5 & H100 \\
Base Model + CR + LR    & 222 & 88   & 375   & H100 \\
TRAAC                   & 418 & 88   & 583   & H100 \\
\bottomrule
\end{tabular}
\caption{Training time breakdown for \shortname and RL baselines during the first GRPO training step.}
\label{tab:grpo_time}
\end{table}

\subsection{FLOPs}
\label{app:flops}

The majority of the difference between \shortname and other RL-based methods lies in the rollout strategy used. \shortname rollout consists of three stages: (i) \textbf{Generation}: producing the initial reasoning steps; (ii) \textbf{Attention-based compression}: computing attention scores for each reasoning step and compressing the trajectory; (iii) \textbf{Answer generation}: generating the final answer based on the compressed chain of thought. Table~\ref{tab:flops_comparison} compares the FLOPs required to generate 20 training examples.

\begin{table}[!h]

\centering
\begin{tabular}{lc}
\toprule
\textbf{Method} & \textbf{FLOPs Used} \\
\midrule
Base Model + CR & $1.65 \times 10^{15}$ FLOPs \\
TRAAC           & $3.84 \times 10^{15}$ FLOPs \\
\bottomrule
\end{tabular}
\caption{FLOPs comparison for generating 20 training examples using different rollout strategies.}
\label{tab:flops_comparison}
\end{table}

Most RL baselines only perform the initial generation step. In contrast, \shortname adds an additional attention-computation stage, yet keeps the overall FLOPs in the same order of magnitude—while producing higher-quality reasoning trajectories. Even though \shortname incurs an increase in training time and FLOPs for the initial batches, mainly due to its multi-stage generation, the overhead remains modest. Moreover, during inference, \shortname makes the model more efficient, effectively reducing the computational cost at test time.

\subsection{Cost amortizes as training progresses}
\label{app:progress_cost}
\begin{figure}[h!]
    \centering
    \includegraphics[width=0.65\linewidth]{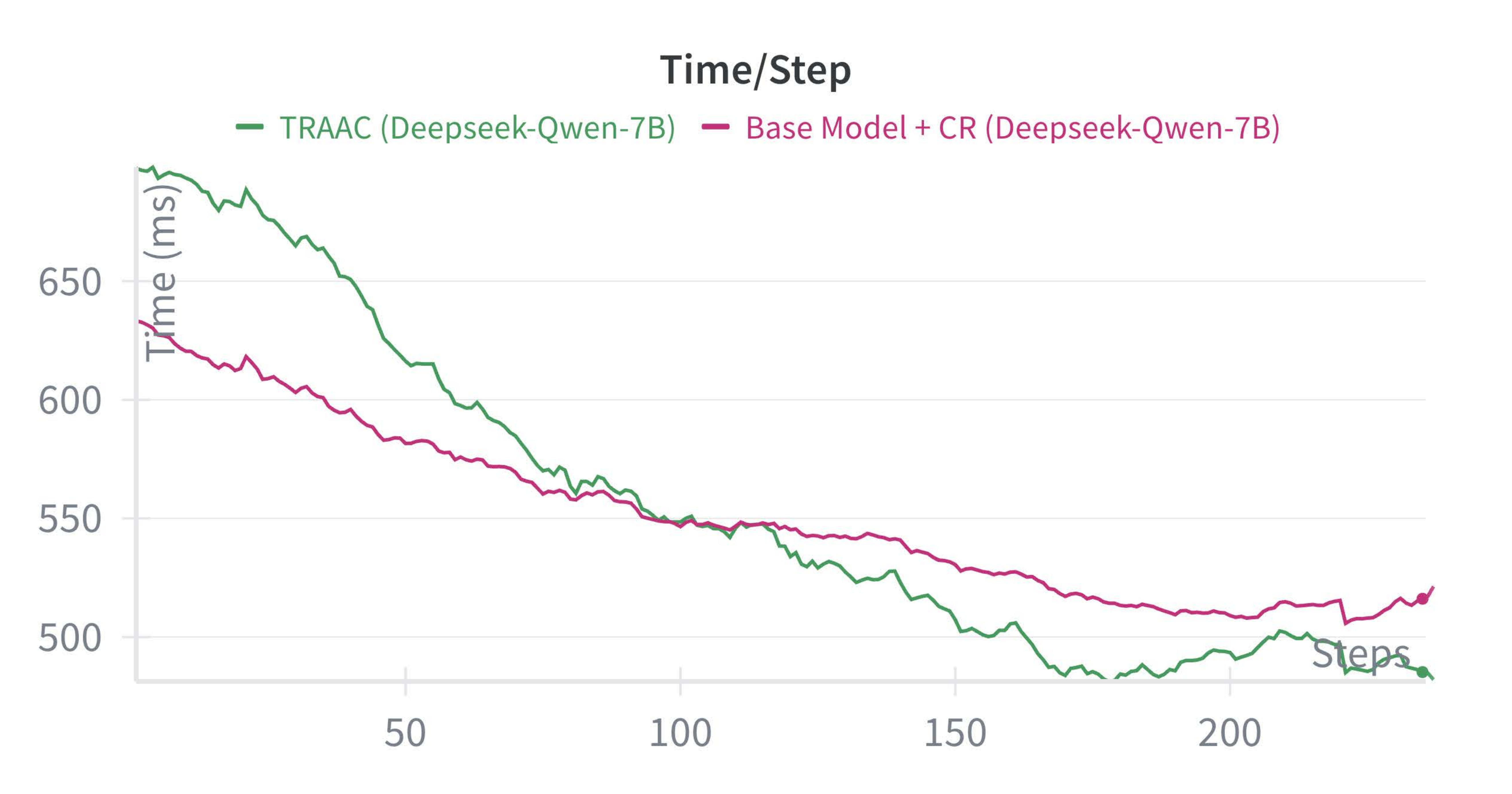}
    \caption{Time per step across training (Deepseek-Qwen-7B)}
    \label{fig:timePerStep}
\end{figure}

The additional overhead introduced at the beginning of the training quickly amortises as the training progresses. As the model learns to shorten its generated reasoning traces, its computational cost – including both FLOPs and time per step steadily decreases. To show this empirically, Figure~\ref{fig:timePerStep} shows the time-per-step curve for training DeepSeek-Qwen-7B. As illustrated in the figure, \shortname begins with a higher step time compared to the Base Model + CR baseline, but the gap closes rapidly. Around mid-training, the two curves match closely, and in later steps, \shortname consistently becomes more efficient – ultimately achieving a lower step time than the baseline. This confirms that while \shortname introduces an initial overhead, its adaptive reduction of reasoning length leads to substantial efficiency gains, resulting in lower computation over the course of training.

To highlight the advantage gained from training TRAAC in terms of FLOPs we calculated the average FLOPs required during inference. We calculated the total amount of FLOPs for generating 80 examples from AMC dataset. The table~\ref{tab:inferenceFlops} show the total amount of FLOPs required for both TRAAC and Base Model + CR.

\begin{table}[h!]

\centering
\small
\begin{tabular}{l c c}
\hline
\textbf{Method} & \textbf{Total FLOPs (80 questions)} & \textbf{Average FLOPs per question} \\
\hline
Base Model + CR (Qwen3-4B) & $3.7 \times 10^{15}$ FLOPs & $4.6 \times 10^{13}$ FLOPs \\
TRAAC (Qwen3-4B)           & $2.7 \times 10^{15}$ FLOPs & $3.3 \times 10^{13}$ FLOPs \\
\hline
\end{tabular}
\caption{Inference compute comparison for TRAAC vs.\ Base Model + CR on 80 AMC questions.}
\label{tab:inferenceFlops}

\end{table}

TRAAC yields a substantial reduction in inference compute. As shown above, TRAAC requires $2.7 \times 10^{15}$ FLOPs to answer 80 AMC questions, compared to $3.7 \times 10^{15}$ FLOPs for the \textit{Base Model + CR}. This corresponds to a $27.3\%$ reduction in inference FLOPs, or a savings of $1.2664 \times 10^{13}$ FLOPs per question. Although TRAAC incurs additional compute at the beginning of training, this overhead amortizes quickly as training progresses. This cost is balanced rapidly at inference time.

Therefore, with the above two experiments, we show that TRAAC not only amortizes quickly during inference but also makes training more efficient within the first 100 steps.

\section{Scaling \shortname to larger response length}

We evaluate scalability of \shortname by increasing the training and test-time token budget from 10k to 15k. As shown in ~\cref{tab:highResponseLength}, across \aime, \amc, and \gpqa datasets, \shortname maintains consistent improvements, achieving an average accuracy gain of 3.5\% and a 23.4\% efficiency gain over \qwen, demonstrating robustness to longer response lengths.
\begin{table}{

\centering
\small
\begin{tabular}{l|ccc}
\toprule
         & \aime                 & \amc                  & \gpqa                 \\ \midrule
Qwen3-4B & 47.74 / 12.3         & 77.11 / 8.5          & 49.64 / 8.6          \\
\shortname     & \textbf{51.93} / 9.7 & \textbf{81.68} / 6.6 & \textbf{51.27} / 6.2 \\ \bottomrule
\end{tabular}
\caption{\shortname with 15k training and test-time response length. 
For each dataset, Accuracy (\%) and Response Length (in $\times$1000 tokens) are reported.}
\label{tab:highResponseLength}}
\end{table}

\section{Attention-based compression identifies redundant steps effectively} 
\label{app:attention}
To help understand the efficiency of the adaptive, attentive compression module, we replace the attention-based compression with random step compression or confidence-based compression. At each training step, instead of using attention as a metric, reasoning steps are pruned either randomly or steps with the least confidence (complete details on how confidence is calculated are in Appendix~\ref{app:confidence}). \cref{tab:compression_ablation} compares \shortname (\qwen) with random steps and least confidence. Relative to \shortname, random step pruning shows an average of 11\% accuracy drop, and similarly, pruning the least confidence steps leads to a 7.25\% accuracy drop. This highlights the efficacy of using attention-based compression in \shortname. Furthermore, to evaluate whether pruned reasoning trajectories maintain their logical coherence when compared to the original reasoning trajectory, we use LLM as a judge to evaluate the logical coherence and flow between the original well formed trajectories and their corresponding pruned version. Results show that \shortname's attention based pruned achieve a higher coherence score when compared with random pruning (complete detail of results in Appendix~\ref{app:coherence}).

\begin{table}

\small
\centering
\begin{tabular}{l|ccc}
\toprule
                                           \bf{Pruning Strategy}    & \multicolumn{1}{l}{\bf\aime} & \multicolumn{1}{l}{\bf\amc} & \multicolumn{1}{l}{\bf\gpqa} \\ \midrule
Random Steps &  29.54 / 6.5                        &    66.74 / 4.1                     &     42.94 / 3.2                     \\
Least Confidence &  32.35 / 5.8                        &    71.08 / 3.4                     &     47 / 3.0                     \\
\shortname     & \textbf{45.45} / 6.7 & \textbf{79.52} / 4.2 & \textbf{47.2} / 4.2 \\ \bottomrule
\end{tabular}
\caption{Ablation on \qwen: comparing \shortname with pruning random and least confident steps. For each dataset, Accuracy(\%) / Response length (k) is reported.}
\label{tab:compression_ablation}
\end{table}

\section{Coherence of pruned reasoning based on attention score}
\label{app:coherence}
To evaluate whether the pruned reasoning trajectories maintain its logical coherence when compared to the original reasoning trajectory, we use LLM as a judge to evaluate the logical coherence and flow between the original well-formed trajectories and their corresponding pruned verion. We use Llama-3.3-70B-Instruct as a judge to evaluate logical coherence between two different pairs of trajectories. Evaluation was done on 200 pairs of trajectories and was assigned a score between 1 to 5, \cref{tab:coherence} reports the logical coherence scores, where each row corresponds to the coherence between a pruned trajectory and its original well-formed (unpruned) version. For pruned trajectories, we use TRAAC’s attention-based trajectories and randomly pruned trajectories. Random pruning provides a controlled baseline that evaluates TRAAC’s attention-based pruning to isolate redundant reasoning steps. Results show an increase of 0.36 when comparing TRAAC’s attention-based pruning with random pruning, which highlights that attention is able to correctly identify which reasoning steps are redundant and effectively prune them.

\begin{table}[!h]

\centering
\begin{tabular}{lc}
\toprule
\textbf{Method} & \textbf{Avg. coherence score} \\
\midrule
Randomly pruned trajectories vs. Original Trajectories & 3.35 \\
TRAAC’s attention-based trajectories vs. Original Trajectories           & 3.71
 \\
\bottomrule
\end{tabular}
\caption{Coherence comparison of attention based pruning strategy vs randomly pruned trajectory.}
\label{tab:coherence}
\end{table}

\section{Agentic Evaluation of \shortname}
\label{app:agentic_Eval}
Including diverse test benchmarks allows us to robustly assess the out-of-distribution generalization capability of TRAAC. In addition, we conduct an evaluation on the agentic, multi-turn benchmark \textbf{MINT}~\citep{wang2024mintevaluatingllmsmultiturn}, which measures an LLM’s ability to solve complex tasks through multi-step interactions and tool use. In MINT, LLMs are tasked with solving problems under different interaction limits $k \in \{1,2,3,4,5\}$, without natural-language feedback. Performance is measured through: (1) the absolute success rate (SR), and (2) the average response length. Table~\ref{tab:mint} compares the base model (Qwen3-4B) and TRAAC across these metrics for three task categories: code generation, decision making, and reasoning.

\begin{table}[h!]

\centering
\small
\resizebox{\textwidth}{!}{
\begin{tabular}{llcccccccccc}
\toprule
\textbf{Task} & \textbf{Method} &
\multicolumn{2}{c}{$k=1$} &
\multicolumn{2}{c}{$k=2$} &
\multicolumn{2}{c}{$k=3$} &
\multicolumn{2}{c}{$k=4$} &
\multicolumn{2}{c}{$k=5$} \\
& &
SR & Len &
SR & Len &
SR & Len &
SR & Len &
SR & Len \\
\midrule

\multirow{2}{*}{code\_generation}
& Base Model
& 0.74 & 0.5
& 58.09 & 3.7
& 58.09 & 4.8
& 59.56 & 5.7
& 59.56 & 7.1 \\

& TRAAC
& 49.26 & 1.7
& 58.82 & 2.6
& 56.62 & 3.3
& 59.56 & 3.5
& 58.82 & 3.8 \\
\midrule

\multirow{2}{*}{decision\_making}
& Base Model
& 0.00 & 0.5
& 11.19 & 2.4
& 17.16 & 2.5
& 30.60 & 2.5
& 33.58 & 3.0 \\

& TRAAC
& 0.00 & 0.5
& 8.21 & 1.0
& 21.64 & 1.2
& 35.07 & 1.3
& 40.30 & 1.6 \\
\midrule

\multirow{2}{*}{reasoning}
& Base Model
& 19.94 & 0.5
& 76.58 & 1.9
& 80.38 & 2.2
& 79.75 & 2.5
& 79.75 & 2.4 \\

& TRAAC
& 66.46 & 0.8
& 76.90 & 1.1
& 81.65 & 1.2
& 79.11 & 1.3
& 81.96 & 1.2 \\
\midrule

\multirow{2}{*}{avg\_micro}
& Base Model
& 10.92 & 0.5
& 57.34 & 2.4
& 60.75 & 2.9
& 63.82 & 3.2
& 64.51 & 3.6 \\

& TRAAC
& 47.27 & 1.0
& 57.00 & 1.4
& 62.12 & 1.7
& 64.51 & 1.8
& 67.06 & 1.9 \\
\bottomrule
\end{tabular}
}
\caption{MINT benchmark results for Base Model (Qwen3-4B) and TRAAC across interaction limits $k \in \{1,2,3,4,5\}$. Metrics include success rate (SR, \%) and average response length.}
\label{tab:mint}
\end{table}

When examining the average micro-aggregated performance across all interaction limits, TRAAC consistently matches or exceeds the base model. TRAAC improves the average success rate by 8.12\% while simultaneously reducing response length by 38.3\%. This demonstrates that TRAAC not only strengthens performance on agentic, multi-turn tasks but also makes the model substantially more efficient in its interactions.

\section{Why choose RL in \shortname}
\label{app:why_rl}
Choosing reinforcement learning (RL) rather than supervised fine-tuning (SFT) to teach adaptive compression is motivated by two key reasons:

\begin{itemize}
    \item \textbf{Generalization advantages of RL.}  
    Prior work has repeatedly shown that RL-based methods yield significantly stronger generalization compared to SFT~\citep{chu2025sftmemorizesrlgeneralizes}. In our own experiments, the SFT-based baseline TokenSkip performs substantially worse than TRAAC, demonstrating that simply applying SFT on compressed outputs is insufficient.

    \item \textbf{Adaptive compression requires an online difficulty signal.}  
    TRAAC relies on dynamically compressing reasoning trajectories according to the difficulty of each problem, where the difficulty signal itself is tightly coupled to the model’s evolving capabilities. Because TRAAC learns this adaptivity during training, an online RL setting naturally allows difficulty estimates to improve alongside the model, enabling progressively better compression decisions.
\end{itemize}

\subsection*{SFT on Attention-Based Compressed Trajectories}

For a direct comparison, we also train an SFT model using reasoning trajectories compressed via \shortname’s attention-based rollout. To generate the dataset, we use a larger model (Qwen3-32B) to produce 1.4k compressed rollouts, and then train a smaller model (Qwen3-4B) on this data. \cref{tab:sft_vs_traac} reports the performance of the SFT model relative to the base model and TRAAC across three benchmarks: AIME, AMC, and GPQA. The SFT model matches the base model’s accuracy while offering moderate efficiency gains. In contrast, TRAAC improves both accuracy and efficiency across all benchmarks.

\begin{table}[h!]

\centering
\small
\begin{tabular}{lcccccc}
\toprule
\textbf{Method} &
\multicolumn{2}{c}{\textbf{AIME}} &
\multicolumn{2}{c}{\textbf{AMC}} &
\multicolumn{2}{c}{\textbf{GPQA}} \\
& Acc. & Len. & Acc. & Len. & Acc. & Len. \\
\midrule
TokenSkip & 5.84\% & 9.6k & 27.71\% & 8.7k & 32.32\% & 7.8k \\
Base Model & 27.64\% & 9.2k & 68.19\% & 7.0k & 45.18\% & 7.6k \\
SFT & 26.06\% & 8.8k & 59.51\% & 6.6k & 42.00\% & 6.9k \\
TRAAC & \textbf{45.45\%} & \textbf{6.7k} & \textbf{79.52\%} & \textbf{4.2k} & \textbf{47.21\%} & \textbf{4.2k} \\
\bottomrule
\end{tabular}
\caption{Comparison of TokenSkip, Base Model, SFT on attention-compressed rollouts, and TRAAC across AIME, AMC, and GPQA. Metrics include accuracy (\%) and average response length.}
\label{tab:sft_vs_traac}
\end{table}

\section{Confidence based compression}
\label{app:confidence}
Similar to attention compression, where a score is calculated for each reasoning token, the confidence of the model is used to calculate the score, and based on the lowest average score, reasoning steps are removed. \cref{alg:token-confidence} shows the pseudocode used to calculate the confidence of each token.

\begin{algorithm}
\begin{algorithmic}[1]

\Require Top-$k$ token log-probabilities
        $L = \{\ell_1, \ell_2, \dots, \ell_k\}$
\Ensure Confidence score $C$

\State \Comment{Convert log-probabilities to probabilities}
\For{$j = 1$ to $k$}
    \State $p_j \gets \exp(\ell_j)$
\EndFor

\State \Comment{Normalize probabilities}
\State $Z \gets \sum_{j=1}^{k} p_j$
\For{$j = 1$ to $k$}
    \State $p_j \gets p_j / Z$
\EndFor

\State \Comment{Compute entropy}
\State $H \gets - \sum_{j=1}^{k} p_j \log(p_j + \epsilon)$

\State \Comment{Maximum entropy with $k$ tokens}
\State $H_{\max} \gets \log(k)$

\State \Comment{Confidence is inverse normalized entropy}
\State $C \gets 1 - (H / H_{\max})$

\State \Return $C$

\end{algorithmic}
\caption{Token Confidence Calculation}
\label{alg:token-confidence}
\end{algorithm}

\section{Attention score over full attention score}

During generation from a reasoning model, the \texttt{</think>} token marks the end of the reasoning process, and the answer tokens generated after \texttt{</think>} contain key summarized conclusions. Prior work~\citep{choi2025thinkclearlyimprovingreasoning} shows that \texttt{</think>} strongly attends to the critical reasoning steps needed to derive the final answer. To verify that attention score from \texttt{</think>} is more effective in practice, table~\ref{tab:full_context} presents additional ablation that computes attention over the full context, including all reasoning tokens as well as the answer tokens generated after \texttt{</think>}. Using these attention scores, low-scoring steps were removed from the reasoning trajectory. The table below compares performance between using attention over the full rollout and using TRAAC.

\begin{table}[h!]
\centering
\begin{tabular}{l|cc|cc|cc}
\toprule
\textbf{Method} & \multicolumn{2}{c|}{\textbf{AIME}} & \multicolumn{2}{c|}{\textbf{AMC}} & \multicolumn{2}{c}{\textbf{GPQA}} \\
 & Acc. & Len. & Acc. & Len. & Acc. & Len. \\
\midrule
TRAAC (full attention rollout) & 2.309\% & 1.9k & 18.53\% & 2.1k & 25.81\% & 4.2k \\
TRAAC & 45.45\% & 6.7k & 79.52\% & 4.2k & 47.21\% & 4.2k \\
\bottomrule
\end{tabular}
\caption{Comparison of TRAAC with full-rollout attention pruning vs.\ standard TRAAC.}
\label{tab:full_context}
\end{table}

Accuracy drops sharply across all three datasets when attention is computed over the full rollout. This demonstrates that without relying on the \texttt{</think>} token for attention scoring, the model cannot reliably identify and prune redundant reasoning steps. Especially on AIME and AMC, we observe a substantial drop in efficiency, indicating that when attention is computed over the complete rollout -- including both reasoning and final answers -- the model struggles to determine which steps are informative versus unnecessary.

\section{\shortname on diverse architecture - Microsoft/Phi-4-mini-reasoning}
To show TRAAC’s robustness across different model architecture, we additionally trained TRAAC on Microsoft/Phi-4-mini-reasoning model, (4B model). \cref{tab:phi_result} presents results comparing the TRAAC-trained Phi-4-mini-reasoning model with the (i) Base Model; (ii): Base Model + CR: The base model trained with GRPO using only the correctness reward (iii) Base model + CR + LR: The base model trained with GRPO using both correctness and length rewards, but without difficulty-level calibration. Overall, \shortname (Phi-4-mini-reasoning) across \aime, \amc and \gpqa achieves an performance improvement of 1.3\% and delivers a 17.5\% boost in efficiency relative to the base model.

\begin{table}[!t]
\centering

{\small
\begin{tabular}{lcccccccc}
\toprule
\multicolumn{1}{l|}{\textbf{Method}}           & \multicolumn{2}{c|}{\textbf{\aime}}     & \multicolumn{2}{c|}{\textbf{\amc}}      & \multicolumn{2}{c|}{\textbf{\gpqa}}     & \multicolumn{2}{c}{\textbf{Average}} \\ \midrule
\multicolumn{1}{l|}{}                          & Acc.$\uparrow$ & \multicolumn{1}{c|}{Len.$\downarrow$} & Acc.$\uparrow$ & \multicolumn{1}{c|}{Len.$\downarrow$} & Acc.$\uparrow$ & \multicolumn{1}{c|}{Len.$\downarrow$} & Acc.$\uparrow$   & Len.$\downarrow$  \\ \midrule
\multicolumn{9}{c}{\textbf{Phi-4-mini-reasoning}}                                                                                                                                                                                                                \\ \midrule
\multicolumn{1}{l|}{Base Model}                & 31.4           & \multicolumn{1}{c|}{8.5}              & 67.4           & \multicolumn{1}{c|}{5.9}              & 40.3           & \multicolumn{1}{c|}{7.7}              & 46.4             & 7.4               \\
\multicolumn{1}{l|}{\: + CR}    & 33.7           & \multicolumn{1}{c|}{8.0}              & 66.3           & \multicolumn{1}{c|}{5.7}              & 38.0           & \multicolumn{1}{c|}{7.6}              & 46.0             & 7.1               \\
\multicolumn{1}{l|}{\: + LR}    & 32.5           & \multicolumn{1}{c|}{8.0}              & 66.3  & \multicolumn{1}{c|}{5.5}              & 42.1           & \multicolumn{1}{c|}{7.2}              & 47.0             & 6.9               \\
\multicolumn{1}{l|}{\shortname} & 28.3  & \multicolumn{1}{c|}{7.7}              & 70.1           & \multicolumn{1}{c|}{5.1}              & 44.5  & \multicolumn{1}{c|}{5.6}              & \textbf{47.6}    & \textbf{6.1}      \\ \bottomrule
\end{tabular}
}
\caption{Results of \shortname on Phi-4-mini-reasoning tested across three datasets: \aime, \amc and \gpqa. Each component addition adds to the previous method.}
\label{tab:phi_result}
\end{table}

\section{\shortname reward design achieves the highest performance.}
\label{app:reward_design}

\begin{table}

\centering
\small
\begin{tabular}{l|ccc}
\toprule
         & \aime                 & \amc                  & \gpqa                 \\ \midrule
$\text{\shortname}_{\text{reduced correctness}}$ & 29.96 / 6.2 & 71.32 / 3.9 & 47.7 / 3.6          \\
$\text{\shortname}_{\text{non adaptive}}$  & 5.33 / 0.6  & 34.87 / 0.7 & 29.79 / 0.5 \\ 
\shortname                & 45.45 / 6.7 & 79.52 / 4.2 & 47.21 / 4.2 \\\bottomrule
\end{tabular}
\caption{Reward ablation results comparing different correctness and length reward configurations.
For each dataset, Accuracy (\%) and Response Length (in $\times$1000 tokens) are reported.}
\label{tab:rewardAblation}
\vspace{-10pt}
\end{table}

During \shortname training, we use a correctness reward of $+4/0$, a format reward of $+1/0$, and a length reward of $+2/0$. The length reward is kept \emph{adaptive}: rollouts receive a positive length reward only if the final answer is correct; otherwise, all length rewards are set to zero. This ensures that correctness is always prioritized over efficiency, which is reflected in the larger magnitude of the correctness reward. To study the impact of this reward design, we conduct two additional ablations: (i) \textbf{Reduced correctness reward:} correctness reward lowered to $+1$, with length reward still adaptive to correctness. (ii) \textbf{Non-adaptive length reward:} correctness reward lowered to $+1$, and length reward made independent of final-answer correctness. Table~\ref{tab:rewardAblation} compares these variants with the full \shortname setup across \aime, \amc, and \gpqa. The ablation results show that reducing the correctness reward maintains reasonable performance on \amc and \gpqa but causes a substantial accuracy drop on \aime. Removing the dependency of the length reward on correctness leads the model to exploit the reward by minimizing output length, resulting in severe performance degradation across all datasets.

\section{Ablation on the length reward smoothing}
\label{app:length_reward_smoothing}

The length reward is a smooth relaxation of a simple preference: among correct rollouts, those shorter than the median length are rewarded more than those longer than the median. A hard step function around the median implements this preference directly, but in our initial experiments it led to unstable training (Section~\ref{sec:rewards}). We therefore use the sigmoid bonus described in Section~\ref{sec:rewards}, written here with an explicit logistic temperature $c$:
\[
\beta \;=\; \frac{1}{1 + \exp\!\left(\dfrac{\ell - \mathrm{median}(L)}{c \cdot \mathrm{median}(L)}\right)},
\label{eq:beta_temp}
\]
where $c$ controls how sharply the reward transitions around the median. As $c \rightarrow 0$, the sigmoid collapses back to the unstable hard step. As $c$ grows toward $0.5$, it flattens and loses the median-anchored signal entirely. We set $c = 0.1$ as a balance point that retains a meaningful gradient around the median.

To test the sensitivity of \shortname to this choice, we run two additional ablations on \qwen. In the first, \emph{naive linear reward}, we drop the smoothing term entirely and use $r_{\text{length}} = L_{\text{norm}}$. In the second, \emph{flattened sigmoid}, we retain $r_{\text{length}} = \max(L_{\text{norm}}, \beta)$ but set $c = 0.5$. Table~\ref{tab:length_reward_ablation} reports the results.

\begin{table}[!t]
\small
\centering
\begin{tabular}{l|ccc}
\toprule
\bf Method                    & \multicolumn{1}{l}{\bf \aime} & \multicolumn{1}{l}{\bf \amc} & \multicolumn{1}{l}{\bf \gpqa} \\ \midrule
Base                      & 27.64 / 9.2              & 68.19 / 7.0             & 45.18 / 7.6              \\
Naive Linear Reward       & 25.30 / 6.6              & 64.09 / 4.2             & 39.80 / 3.2              \\
Flattened Sigmoid ($c = 0.5$) & 31.91 / 6.4          & 71.32 / 4.0             & 44.87 / 2.8              \\
\shortname ($c = 0.1$)    & \textbf{45.45} / 6.7     & \textbf{79.52} / 4.2    & \textbf{47.21} / 4.2     \\ \bottomrule
\end{tabular}
\caption{Ablation on the length reward smoothing using \qwen. For each dataset Accuracy(\%) / Response length (k) are reported.}
\label{tab:length_reward_ablation}
\end{table}

Removing the sigmoid entirely drops accuracy below the base model on all three datasets, confirming that the smoothing is necessary for stable training. Flattening the sigmoid to $c = 0.5$ recovers some of this accuracy relative to the linear variant, but still falls short of $c = 0.1$. A flatter $\beta$ makes the reward nearly constant around the median, so rollouts of meaningfully different length receive nearly the same length reward, which removes the signal that drives adaptive length allocation.

\section{Attention-based compression during test time inference.}
\label{app:test_time}

Similar to prior test-time approaches~\citep{choi2025thinkclearlyimprovingreasoning}, we evaluated \shortname as a test-time method by adapting its compression module to operate during inference. Specifically, during decoding, once the model outputs \texttt{</think>}, we apply the attention-based compression module with a static compression rate of 0.4 (same as a medium difficulty task), pruning intermediate reasoning steps. Because task difficulty cannot be estimated at test time, this static compression rate is maintained throughout. If \texttt{</think>} is not produced, no compression is applied. After compression, the model is allowed to generate the final answer. \cref{tab:test_time} presents the results on \qwen, and compares this inference-only variant of \shortname against both the base model and the fully trained \shortname. On average across \aime, \amc and \gpqa datasets, the fully trained \shortname outperforms the inference-only variant by 7.28\% in accuracy and 27.5\% in efficiency, underscoring the benefit of incorporating the compression module during training. On \gpqa, the test-time method achieves accuracy comparable to \shortname but suffers from a 38\% efficiency drop. Finally, even the inference-only setup yields an average accuracy improvement of 3.11\% and 12.65\% efficiency gain over the base model, indicating that applying compression at test time already provides measurable accuracy gains, while training with compression further amplifies accuracy by 10.39\% and efficiency by 36.7\%.

\begin{table}[!h]

\small
\centering
\begin{tabular}{l|ccc}
\toprule
\bf Method                    & \multicolumn{1}{l}{\bf \aime} & \multicolumn{1}{l}{\bf \amc} & \multicolumn{1}{l}{\bf \gpqa} \\ \midrule
Base                      & 27.64 / 9.2              & 68.19 / 7.0             & 45.18 / 7.6              \\
Test-Time                 &        32.13 / 8.5                  &           70.60 / 5.7             &       \textbf{47.61} / 6.6                   \\
\shortname     & \textbf{45.45} / \textbf{6.7} & \textbf{79.52} / \textbf{4.2} & 47.20 / \textbf{4.2} \\ \bottomrule
\end{tabular}
\caption{Results of \shortname as a test time method using \qwen, compared to base model and \shortname. For each dataset Accuracy(\%) / Response length (k) are reported.}
\label{tab:test_time}
\end{table}

    \section{Deepseek Ablation and Analysis}
    \cref{tab:deepseek_ablation} and \cref{tab:deepseek_optimal_ablation} present the ablation results for \textbf{(i) Base Model + CR:} The base model trained with GRPO using only the correctness reward, \textbf{(ii) Base model + CR + LR:} The base model trained with GRPO using both correctness and length rewards, but without difficulty-level calibration.
\label{appendix:deepseek_analysis}
\begin{table}[!h]
\centering

{\small
\begin{tabular}{lcccccccccc}
\toprule
\multicolumn{1}{l|}{\textbf{Method}}           & \multicolumn{2}{c|}{\textbf{\aime}}         & \multicolumn{2}{c|}{\textbf{\amc}}          & \multicolumn{2}{c|}{\textbf{\gpqa}}         & \multicolumn{2}{c|}{\textbf{\bbeh}}         & \multicolumn{2}{c}{\textbf{Average}} \\ \midrule
\multicolumn{1}{l|}{}                          & Acc.$\uparrow$           & \multicolumn{1}{c|}{Len.$\downarrow$} & Acc.$\uparrow$           & \multicolumn{1}{c|}{Len.$\downarrow$} & Acc.$\uparrow$           & \multicolumn{1}{c|}{Len.$\downarrow$} & Acc.$\uparrow$           & \multicolumn{1}{c|}{Len.$\downarrow$} & Acc.$\uparrow$                  & Len.$\downarrow$         \\ \midrule
\multicolumn{11}{c}{\textbf{DeepSeek-R1-Distill-Qwen-7B}}                                                                                                                                                                                                     \\ \midrule
\multicolumn{1}{l|}{Base Model}                & 33.71          & \multicolumn{1}{c|}{8.2}  & 74.22          & \multicolumn{1}{c|}{5.7}  & 43.55          & \multicolumn{1}{c|}{7.1}  & 10.61          & \multicolumn{1}{c|}{5.9}  & 40.5                  & 6.7          \\
\multicolumn{1}{l|}{\: + CR}           & 35.81          & \multicolumn{1}{c|}{7.6}  & 78.55          & \multicolumn{1}{c|}{4.9}  & 45.99          & \multicolumn{1}{c|}{6.1}  & \textbf{11.74} & \multicolumn{1}{c|}{5.1}  & 43.0                  & 5.9          \\
\multicolumn{1}{l|}{\: + LR}      & 32.73          & \multicolumn{1}{c|}{6.0}  & \textbf{79.04} & \multicolumn{1}{c|}{3.3}  & 45.99          & \multicolumn{1}{c|}{3.5}  & 11.51          & \multicolumn{1}{c|}{2.7}  & 42.3                  & 3.9          \\
\multicolumn{1}{l|}{\shortname} & \textbf{38.60} & \multicolumn{1}{c|}{7.3}  & 77.83          & \multicolumn{1}{c|}{4.5}  & \textbf{47.31} & \multicolumn{1}{c|}{6.2}  & 11.55          & \multicolumn{1}{c|}{5.2}  & \textbf{43.8}         & 5.8          \\ \bottomrule
\end{tabular}
}
\caption{Ablation Results of \shortname on \deepseek tested across 4 datasets: \aime, \amc, \gpqa, and \bbeh. Each component addition adds to the previous method.}
\label{tab:deepseek_ablation}
\end{table}

\begin{table}[!h]
\centering

{\small
\begin{tabular}{lcccccc}
\toprule
\multicolumn{1}{l|}{\textbf{Method}}           & \multicolumn{3}{c|}{\textbf{\overthinkBench}}                     & \multicolumn{2}{c|}{\textbf{\underthinkBench}}   & \textbf{\optimalThinkBench} \\ \midrule
\multicolumn{1}{l|}{}                          & Acc.$\uparrow$           & Len.$\downarrow$ & \multicolumn{1}{c|}{$\text{AUC}_{\text{OAA}}\uparrow$}            & Acc.$\uparrow$           & \multicolumn{1}{c|}{Len.$\downarrow$} & F1$\uparrow$                            \\ \midrule
\multicolumn{7}{c}{\textbf{DeepSeek-R1-Distill-Qwen-7B}}                                                                                                                      \\ \midrule
\multicolumn{1}{l|}{Base Model}                & 78.45          & 0.9  & \multicolumn{1}{c|}{72.38}          & 12.69          & \multicolumn{1}{c|}{6.2}  & 21.6                          \\
\multicolumn{1}{l|}{\: + CR}           & 79.51          & 0.8  & \multicolumn{1}{c|}{\textbf{73.36}} & 17.05          & \multicolumn{1}{c|}{5.7}  & 27.7                          \\
\multicolumn{1}{l|}{\: + LR}      & 78.06          & 0.4  & \multicolumn{1}{c|}{72.61}          & 14.69          & \multicolumn{1}{c|}{3.0}  & 24.4                          \\
\multicolumn{1}{l|}{\shortname} & \textbf{81.81} & 1.0  & \multicolumn{1}{c|}{72.89}          & \textbf{22.30} & \multicolumn{1}{c|}{5.9}  & \textbf{34.1}                 \\ \bottomrule
\end{tabular}
}
\caption{Ablation Results of \shortname (\deepseek) on \optimalThinkBench. Each component addition adds to the previous method.}
\label{tab:deepseek_optimal_ablation}
\end{table}

Note that \deepseek reported number 33.71\% on \aime are different from ~\cite{deepseekai2025deepseekr1incentivizingreasoningcapability} because they , using a generation length of 32k tokens and 64 sampled responses per query to compute pass@1 accuracy. However, our evaluation setup is different, we evaluate models on all questions combined from AIME 22, AIME 23, and AIME 24, and use a maximum generation length of 10k tokens and report the mean accuracy over 5 runs. Therefore there is a difference between the reported numbers. To verify the correctness of our evaluation script, we replicated the exact setup used in original work and obtained an accuracy of 53.8\%, which is very close to the reported value of 55.5\% in \cite{deepseekai2025deepseekr1incentivizingreasoningcapability}.

\section{Compression Module}
\label{appendix:comprresion_module}
\subsection{Prompt}
\label{appendix:compression_prompt}
For every reasoning trajectory, auxiliary prompt was appended at the end of the trajectory. The prompt is: ``Time is up. I should stop thinking and now write a summary containing all key steps required to solve the problem.''.
\subsection{Special Tokens to split Trajectory to Chunks}
\label{app:special_tokens}
Below is the list that is used to split each reasoning trajectory into multiple reasoning steps.
\begin{verbatim}
split_tokens = [
    "Wait", "Alternatively", "Another angle", "Another approach", "But wait",
    "Hold on", "Hmm", "Maybe", "Looking back", "Okay", "Let me", "First",
    "Then", "Alright", "Compute", "Correct", "Good", "Got it",
    "I don't see any errors", "I think", "Let me double-check", "Let's see",
    "Now", "Remember", "Seems solid", "Similarly", "So", "Starting",
    "That's correct", "That seems right", "Therefore", "Thus"
]
\end{verbatim}
\subsection{Uniformity Score}
\label{appendix:uniformity_score}
\cref{alg:eviction} presents the pseudocode for calculating the uniformity score, based on which the final compression rate is calculated.

\begin{algorithm}[tb]
\caption{Calculating Eviction Percentage Based on Attention Uniformity}
\label{alg:eviction}
\begin{algorithmic}[1]

\Require Step importance scores $\{s_1, s_2, \dots, s_n\}$,
        target reduction $\tau$ (default: $0.25$)
\Ensure Eviction percentage $e \in [0,1]$

\Function{CalculateUniformityScore}{$\{s_1, \dots, s_n\}$}
    \If{$n \le 1$}
        \State \Return $1.0$ \Comment{Single step $\Rightarrow$ uniform}
    \EndIf
    \State Clamp all $s_i \ge 0$
    \State $T \gets \sum_i s_i$
    \If{$T \le 0$}
        \State \Return $1.0$
    \EndIf
    \State $p_i \gets s_i / T$
    \State $H \gets -\sum_i p_i \log(p_i + \epsilon)$ \Comment{$\epsilon = 10^{-12}$}
    \State $H_{\max} \gets \log(n)$
    \If{$H_{\max} = 0$}
        \State \Return $1.0$
    \EndIf
    \State \Return $H / H_{\max}$
\EndFunction

\Function{DetermineEvictionPercentage}{$u, \tau$}
    \If{$u > 0.8$}
        \State \Return $0.0$ \Comment{High uniformity}
    \EndIf
    \State $e \gets \tau (1 - u)$
    \State \Return $\min(e, 0.8)$
\EndFunction

\State $u \gets$ \Call{CalculateUniformityScore}{$\{s_1, \dots, s_n\}$}
\State $e \gets$ \Call{DetermineEvictionPercentage}{$u, \tau$}

\end{algorithmic}
\end{algorithm}

\section{GRPO Details}
\label{appendix:grpo}
For each question $q$, a group of responses $\{y^{1}, y^{2}, \ldots, y^{N}\}$ is sampled from the old policy $\pi_{\text{old}}$, and the policy model $\pi_{\theta}$ is optimized by maximizing the following GRPO objective.

\begin{equation}
\mathcal{J}_{\text{GRPO}}(\theta) = \frac{1}{N} \sum_{i=1}^{N} \frac{1}{|y^i|} \sum_{t=1}^{|y^i|}
\min \left[
\frac{\pi_\theta(y^i(t) | y^i_{<t})}{\pi_{\text{old}}(y^i(t) | y^i_{<t})} \hat{A}_{i,t},\,
\text{clip} \left(
\frac{\pi_\theta(y^i(t) | y^i_{<t})}{\pi_{\text{old}}(y^i(t) | y^i_{<t})}, 1 - \varepsilon, 1 + \varepsilon
\right) \hat{A}_{i,t}
\right], \notag
\end{equation}

where $\varepsilon$ is the clipping range hyperparameter, and $\hat{A}_{i,t}$ represents the advantage, 
computed based on the relative verifiable outcome based rewards of outputs within each group. Both the original and the compressed rollouts remain in distribution with respect to the current policy. The original rollouts are sampled directly from $\pi_{\text{old}}$, and the compressed rollouts are constructed by pruning the policy's own generation and regenerating the final answer from that same policy. Because compressed trajectories are derived from on-policy samples rather than drawn from an external distribution, they remain approximately in distribution, and the objective above stays well-defined for every rollout in the group.

\section{Experimental Details}
\label{app:experimental_details}
We adopt verl~\citep{sheng2024hybridflow} as the training framework.
\subsection{Hyperparameters}
\label{appendix:hyperparameters}

\begin{table}[!h]
\centering
\begin{tabular}{l c l}
\toprule
\textbf{Category} & \textbf{Hyperparameter} & \textbf{Value} \\
\midrule
\multirow{9}{*}{Training} 
    & Number of rollouts    & 8 \\
    & Temperature           & 1.0 \\
    & top\_p                & 1.0 \\
    & top\_k                & -1.0 \\
    & Max response length   & 10k \\
    & clip\_ratio\_low      & 0.20 \\
    & clip\_ratio\_high     & 0.28 \\
    & kl\_loss\_coef        & 0.001 \\
    & Learning rate (LR)    & 1e-6 \\
\midrule
\multirow{6}{*}{Evaluation} 
    & Number of rollouts    & 8 \\
    & Temperature           & 1.0 \\
    & top\_p                & 1.0 \\
    & top\_k                & -1.0 \\
    & Max response length   & 10k \\
    & N                     & 5 \\
\midrule
\multirow{3}{*}{Difficulty Calibration} 
    & Hard                  & 0.20 \\
    & Medium                & 0.40 \\
    & Easy                  & 0.60 \\
\bottomrule
\end{tabular}
\caption{Hyperparameters used for training, evaluation, and difficulty calibration.}
\end{table}

\subsection{Training Reward}
\label{app:training_reward}
To ensure a high weight on correctness relative to other components, we assign a 
\textbf{correctness reward} of $+4$ if the final answer is correct and $0$ otherwise. 
The \textbf{format reward} ranges from $0$ to $1$: a score of $0.5$ is given for the 
presence of the \texttt{<think>} and \texttt{</think>} tokens, and an additional $0.5$ 
is awarded if every reasoning trajectory is properly enclosed within these tokens in 
the correct order. The \textbf{length reward} ranges from $0$ to $2$. The overall reward 
is computed as the sum of these components:
\[
\text{Total Reward} = \text{Correctness Reward} + \text{Format Reward} + \text{Length Reward}.
\]

\subsection{Evaluation Metrics}
\label{appendix:eval_metrics}
For each of the dataset we compute the accuracy and the average response length. Specifically for \overthinkBench we also compute the $\text{AUC}_\text{OAA}$. This metric is based on Overthinking-Adjusted Accuracy (OAA), which measures model correctness under a limit on reasoning tokens. 
For a threshold $t$, it is defined as
\[
\text{OAA}_t = \frac{1}{n} \sum_{i=1}^{n} 
\big( \text{Correctness}_i \cdot \mathbb{I}(\text{ThinkTokens}_i < t) \big),
\]
where $\text{Correctness}_i \in \{0,1\}$ indicates whether the $i$-th response is correct, and $\mathbb{I}(\cdot)$ is the indicator function that enforces the thinking length constraint.

\[
\text{AUC}_{\text{OAA}} = \int_{0}^{t_{\max}} \frac{\text{OAA}_t}{t_{\max}} \, dt 
    \approx \frac{1}{t_{\max}} \sum_{t=0}^{t_{\max}} \text{OAA}_t,
\]
where $t_{\max}$ is the maximum number of allowed thinking tokens. Furthermore, following the method from ~\citep{aggarwal2025optimalthinkingbenchevaluatingunderthinkingllms}, to compute the \optimalThinkBench metric: F1 score we combine the  $\text{AUC}_\text{OAA}$ from \overthinkBench and Accuracy ($\mathrm{Acc}_{\mathrm{ut}}$) from \underthinkBench into a single F1 score:
\begin{equation}
F1 = 2 \cdot \frac{\mathrm{AUC}_{\mathrm{OAA}} \cdot \mathrm{Acc}_{\mathrm{ut}}}{\mathrm{AUC}_{\mathrm{OAA}} + \mathrm{Acc}_{\mathrm{ut}}}
\end{equation}

\subsection{Training prompt}
For each questions in the training set, instruction was provided: \texttt{``Let's think step by step and output the final answer within $\backslash \backslash \text{\texttt{boxed}}\{\}$''}

\subsection{Baseline prompts}
\label{app:Baseline_prompt}
Below we define the instruction that was provided to each baseline model, each baseline follows the prompt officially used in the original work:
\begin{itemize}
    \item Base Model: \texttt{``Let's think step by step and output the final answer within $\backslash \backslash \text{\texttt{boxed}}\{\}$''}
    \item L1-Max: \texttt{``Let's think step by step and output the final answer within $\backslash \backslash \text{\texttt{boxed}}\{\}$. Think for maximum 10000 tokens.''}
    \item LC-R1: \texttt{`` Please reason step by step, and put your final answer within $\backslash \backslash \text{\texttt{boxed}}\{\}$''}
    \item AdaptThink: No prompt, just the question
    \item TokenSkip: \texttt{"<|im\_start|>system You are a helpful assistant.<|im\_end|> <|im\_start|>user Please reason step by step, and put your final answer within \textbackslash boxed\{\}. question<|eot\_id|>0.5<|eot\_id|><|im\_end|> <|im\_start|>assistant"}

\end{itemize}
\subsection{Compute Used}
All training was done on 4*A100 (80GB).

\section{Examples}

\subsection{Example 1 (\shortname produces shorter reasoning traces)}

\textbf{Data Source:} AMC

\textbf{Input:}
\begin{tcolorbox}[
  breakable,
  listing only,
  colback=gray!20,
  colframe=gray!30,
  boxrule=0.3pt,
  arc=2pt,
  listing options={basicstyle=\ttfamily\small, breaklines=true}
]
Maureen is keeping track of the mean of her quiz scores this semester. If Maureen scores an $11$ on the next quiz, her mean will increase by $1$. If she scores an $11$ on each of the next three quizzes, her mean will increase by $2$. What is the mean of her quiz scores currently? Let's think step by step and output the final answer within boxed\{\}
\end{tcolorbox}

\textbf{Response (TRAAC):}
\begin{tcolorbox}[
    colback=gray!20,
    colframe=gray!30,
    breakable,
    sharp corners,
    boxrule=0.4pt
]
\begin{verbatim}
<think>\nOkay, let's try to solve this problem. So, Maureen's current mean 
is what we need to find. The problem gives two scenarios. Let's break it down.
First, if she scores an 11 on the next quiz, her mean increases by1. Second, 
if she gets11 on each of the next three quizzes, the mean increases by2. 
Hmm. Let's think about how means work. Let's say currently she has a mean of M. 
Let's say she has n quizzes so far. The total sum of her current scores 
be M multiplied by n. Wait, when she takes another quiz, the number of quizzes 
becomes n+1. For the first scenario, adding an11would make the new mean be 
M +1. So the new total would be (current sum) +11. And that divided by (n+1) 
equals the new mean. So (current sum) +11 = (M+1) * (n+1). But the current 
sum is M *n. So substituting, M*n +11 = (M+1)(n+1). Let's expand the right
side: M*n +M +n +1. So the equation would be M*n +11 = M*n +M +n +1. 
Subtract M*n from both sides:11 = M +n +1. So M +n equals10? Because11-1 
is10. So M +n=10? That's equation one.\n\nThen the second scenario: if she 
scores11 on each of thenext three quizzes. So adding three quizzes, each11. 
The new mean would be current mean plus2. Let's think. The current sum is 
M*n. After three more quizzes, the total sum is M*n +3*11. The number of 
quizzes is n+3. The new mean is M+2. So (M*n +33) / (n+3) = M+2. Let's 
multiply both sides by(n+3): M*n +33 = (M+2)(n+3). Let's expand the right 
side: M*n +3M +2n +6. So the equation becomes M*n +33 = M*n +3M +2n 
+6. Subtract M*n from both sides:33=3M +2n +6. Subtract6:27=3M +2n. So 
that's the second equation.\n\nSo from the first scenario, we had M +n=10. 
Let's call that equation1. And the second gives3M +2n=27? Equation2. So we 
need to solve these two equations. Let's see. From equation1, n is10-M. 
Substitute into equation2. So3M +2*(10-M) =27. Let's calculate that:3M +20-
2M =27. So (3M-2M) is M. So M +20=27. Therefore M is7? Because27-20is7. So 
the current mean is7? Let's check. \n\nIf M is7, then n is10-M which is3. So 
currently, she has3 quizzes. Current total is7*3=21. First scenario: 
adding11. New total is21+11=32. Number of quizzes is4. Mean would be32/4=8. 
Which is7+1. Correct. For the second scenario, adding three11s. The total 
would be21+33=54. Number of quizzes is3+3=6.54/6 is9. The current mean was7, 
so7+2=9. Yes that's right. So the current mean is7?\n</think>\n\nTo 
determine the current mean of Maureen's quiz scores, we analyze the two given 
scenarios mathematically.\n\n---\n\n### Step 1: Define 
Variables\n\nLet:\n- $ M $ = current mean of quiz scores\n- $ n $ = 
number of quizzes Maureen has taken so far\n\nThen the total of her current 
scores is $ M \\times n $.\n\n---\n\n### Step 2: First 
Scenario\n\nIf Maureen scores an 11 on the next quiz, her mean increases by 
1.\n\nThe new total becomes $ M \\times n + 11 $, and the number of 
quizzes is $ n + 1 $. The new mean is $ M + 1 $. So:\n\n$$\n\frac{Mn 
+ 11}{n + 1} = M + 1\n$$\n\nMultiply both sides by $ n + 1 
$:\n\n$$\nMn + 11 = (M + 1)(n + 1)\n$$\n\nExpanding the right-hand 
side:\n\n$$\nMn + 11 = Mn + M + n + 1\n$$\n\nSubtract $ Mn $ from 
both sides:\n\n$$\n11 = M + n + 1 \Rightarrow M + n = 10 \quad 
\text{(Equation 1)}\n$$\n\n---\n\n### Step 3: Second 
Scenario\n\nIf she scores 11 on each of the next three quizzes, the mean 
increases by 2.\n\nThe total becomes $ Mn + 33 $, and the number of quizzes 
becomes $ n + 3 $. The new mean is $ M + 2 $. So:\n\n$$\n\\frac{Mn + 
33}{n + 3} = M + 2\n$$\n\nMultiply both sides by $ n + 3 
$:\n\n$$\nMn + 33 = (M + 2)(n + 3)\n$$\n\nExpanding the right-hand 
side:\n\n$$\nMn + 33 = Mn + 3M + 2n + 6\n$$\n\nSubtract $ Mn $ from 
both sides:\n\n$$\n33 = 3M + 2n + 6 \\Rightarrow 3M + 2n = 27 
\quad \text{(Equation 2)}\n$$\n\n---\n\n### Step 4: Solve the 
System of Equations\n\nFrom Equation 1:\n$$\nM + n = 10 \\Rightarrow n 
= 10 - M\n$$\n\nSubstitute into Equation 2:\n\n$$\n3M + 2(10 - M) = 
27 \Rightarrow 3M + 20 - 2M = 27 \\Rightarrow M + 20 = 27 
\Rightarrow M = 7\n$$\n\n---\n\n### Final 
Answer\n\n$$\n\boxed{7}\n$$
\end{verbatim}
\end{tcolorbox}

\textbf{Response (Base Model + CR):}
\begin{tcolorbox}[
    colback=gray!20,
    colframe=gray!30,
    breakable,
    sharp corners,
    boxrule=0.4pt
]
\begin{verbatim}
<think>\nOkay, let me try to figure out this problem. So, Maureen is 
tracking her mean quiz score. The problem says that if she scores an 11 on the 
next quiz, her mean will increase by 1. Also, if she scores 11 on each of the 
next three quizzes, her mean will increase by 2. We need to find her current 
mean. Hmm, let me break this down step by step.\n\nFirst, let me recall what 
a mean is. The mean is the sum of all the scores divided by the number of 
scores. So, if I let her current mean be something, maybe I can set up 
equations based on the information given.\n\nLet me denote the current 
number of quizzes she has taken as $ n $, and her current total score as $ S 
$. Therefore, her current mean is $ \\frac{S}{n} $.\n\nNow, the first 
condition: if she scores an 11 on the next quiz, her mean will increase by 1. 
Let's think about that. If she takes one more quiz with a score of 11, then 
her total score becomes $ S + 11 $, and the number of quizzes becomes $ n + 
1 $. The new mean is $ \\frac{S + 11}{n + 1} $. According to the problem, 
this new mean is 1 more than her current mean. So, we can write the 
equation:\n\n$$\n\frac{S + 11}{n + 1} = \frac{S}{n} + 
1\n$$\n\nOkay, that seems like a good starting point. Let me write that 
down.\n\nSimilarly, the second condition: if she scores an 11 on each of the 
next three quizzes, her mean will increase by 2. So, taking three more 
quizzes with 11 each, her total score becomes $ S + 11*3 = S + 33 $, and 
the number of quizzes becomes $ n + 3 $. The new mean is $ \frac{S + 33}
{n + 3} $, which should be 2 more than her current mean. So, the equation 
would be:\n\n$$\n\frac{S + 33}{n + 3} = \frac{S}{n} + 
2\n$$\n\nSo, now we have two equations with two variables, $ S $ and $ n $. 
Let me write both equations again:\n\n1. $ \frac{S + 11}{n + 1} = 
\frac{S}{n} + 1 $\n2. $ \frac{S + 33}{n + 3} = \frac{S}{n} + 2 
$\n\nI need to solve these equations for $ S $ and $ n $, and then find the 
current mean $ \frac{S}{n} $.\n\nLet me start with the first equation. 
Let's simplify it.\n\nMultiply both sides by $ n(n + 1) $ to eliminate 
denominators:\n\n$ (S + 11) \cdot n = (S + n(n + 1)) \cdot 1 
$\n\nWait, let me check that again. Wait, the right-hand side is $ 
\frac{S}{n} + 1 $, so when I multiply both sides by $ n(n + 1) $, it 
becomes:\n\nLeft side: $ (S + 11) \cdot n $\n\nRight side: $ [ 
\frac{S}{n} + 1 ] \cdot n(n + 1) $\n\nWait, let me do that more 
carefully:\n\nOriginal equation:\n\n$ \frac{S + 11}{n + 1} = 
\frac{S}{n} + 1 $\n\nMultiply both sides by $ n(n + 1) $:\n\nLeft 
side: $ (S + 11) \cdot n $\n\nRight side: $ [ \frac{S}{n} + 1 ] 
\cdot n(n + 1) = [ S + n ] \cdot (n + 1) $\n\nWait, because $ 
\frac{S}{n} \times n(n + 1) = S(n + 1) $, and $ 1 \times n(n + 
1) = n(n + 1) $. Therefore:\n\nRight side: $ S(n + 1) + n(n + 1) 
$\n\nTherefore, equation becomes:\n\n$ n(S + 11) = S(n + 1) + n(n + 
1) $\n\nLet me expand both sides:\n\nLeft side: $ nS + 11n $\n\nRight 
side: $ Sn + S + n^2 + n $\n\nNow, subtract left side from both 
sides:\n\n0 = (Sn + S + n² + n) - (nS + 11n)\n\nSimplify:\n\nSn + 
S + n² + n - nS -11n = S + n² -10n = 0\n\nTherefore:\n\nS + n² 
-10n = 0 → S = -n² +10n\n\nSo that's equation 1: S = -n² 
+10n\n\nOkay, now let's do the same for the second equation:\n\n$ 
\frac{S + 33}{n + 3} = \frac{S}{n} + 2 $\n\nMultiply both sides by 
$ n(n + 3) $:\n\nLeft side: (S + 33) * n\n\nRight side: [ S/n + 2 
] * n(n + 3) = [ S + 2n ] * (n + 3 )\n\nTherefore:\n\nLeft side: 
nS + 33n\n\nRight side: (S + 2n)(n + 3) = S(n + 3) + 2n(n + 3) = 
Sn + 3S + 2n² + 6n\n\nTherefore, equation:\n\nnS + 33n = Sn + 3S + 
2n² + 6n\n\nSubtract Sn from both sides:\n\n33n = 3S + 2n² 
+6n\n\nBring all terms to left side:\n\n33n -6n -2n² -3S = 0 → 27n 
-2n² -3S = 0\n\nRearranged:\n\n-2n² +27n -3S = 0\n\nMultiply both 
sides by -1:\n\n2n² -27n +3S =0\n\nSo:\n\n3S = -2n² +27n → S = 
(-2n² +27n)/3\n\nSo now, from equation 1, we have S = -n² +10n, and from 
equation 2, S = (-2n² +27n)/3.\n\nTherefore, we can set them equal:\n\n-
n² +10n = (-2n² +27n)/3\n\nMultiply both sides by 3:\n\n-3n² +30n = 
-2n² +27n\n\nBring all terms to left side:\n\n-3n² +30n +2n² -27n = 
0 → (-n²) +3n =0 → -n² +3n =0 → n(-n +3)=0\n\nSo, solutions are n=0 or 
-n +3=0 → n=3.\n\nBut n is the number of quizzes, so it can't be zero. 
Therefore, n=3.\n\nTherefore, the current number of quizzes is 3.\n\nThen, 
substitute n=3 into equation 1: S = -n² +10n = -9 +30 = 
21\n\nTherefore, current total score S=21, and number of quizzes n=3, so 
current mean is 21/3=7.\n\nWait, so the current mean is 7?\n\nBut let me 
check if this works with the original conditions.\n\nFirst condition: if she 
scores an 11 on the next quiz, the mean increases by 1.\n\nCurrent mean is 
7, with 3 quizzes.\n\nAfter taking 11, total is 21 +11=32, number of 
quizzes is 4. New mean is 32/4=8, which is 7 +1=8. Correct.\n\nSecond 
condition: if she scores 11 on each of the next three quizzes, then total is 
21 +33=54, number of quizzes is 3 +3=6. New mean is 54/6=9, which is 7 
+2=9. Correct. Perfect.\n\nTherefore, current mean is 7. So the answer is 
7, which is \boxed{7}\n\n**Final Answer**\nThe mean of Maureen's 
quiz scores currently is \boxed{7}.\n</think>\n\nTo determine 
Maureen's current mean quiz score, we start by modeling the situation 
mathematically.\n\n---\n\n### Step 1: Define Variables\n\nLet:\n- $ 
n $ be the current number of quizzes Maureen has taken.\n- $ S $ be her total 
score so far.\n- Her current mean is then $ \frac{S}{n} $.\n\n---
\n\n### Step 2: Use Given Conditions to Form Equations\n\n#### 
First Condition:\nIf Maureen scores an **11** on the next quiz, her mean 
increases by 1.\n\nNew total score: $ S + 11 $  \nNew number of quizzes: $ 
n + 1 $  \nNew mean: $ \frac{S + 11}{n + 1} $\n\nAccording to the 
problem:\n$$\n\frac{S + 11}{n + 1} = \frac{S}{n} + 
1\n$$\n\nMultiply both sides by $ n(n + 1) $ to eliminate 
denominators:\n$$\n(S + 11)n = S(n + 1) + n(n + 
1)\n$$\n\nSimplify:\n$$\nSn + 11n = Sn + S + n^2 + 
n\n\Rightarrow 11n = S + n^2 + n\n\Rightarrow S = -n^2 + 
10n\n$$\n\n#### Second Condition:\nIf Maureen scores an **11** on 
**three** quizzes, her mean increases by 2.\n\nNew total score: $ S + 
33 $  \nNew number of quizzes: $ n + 3 $  \nNew mean: $ \frac{S + 33}{n 
+ 3} $\n\nAccording to the problem:\n$$\n\frac{S + 33}{n + 3} = 
\frac{S}{n} + 2\n$$\n\nMultiply both sides by $ n(n + 3) $:\n$$\n(S 
+ 33)n = (S + 2n)(n + 3)\n$$\n\nSimplify:\n$$\nSn + 33n = Sn + 3S 
+ 2n^2 + 6n\n\Rightarrow 33n = 3S + 2n^2 + 6n\n\Rightarrow 27n = 
3S + 2n^2\n\Rightarrow 3S = -2n^2 + 27n\n\Rightarrow S = 
\frac{-2n^2 + 27n}{3}\n$$\n\n---\n\n### Step 3: Solve for $ n $ 
and $ S $\n\nFrom the two expressions for $ S $:\n$$\n-n^2 + 10n = 
\frac{-2n^2 + 27n}{3}\n$$\n\nMultiply both sides by 3:\n$$\n-3n^2 + 
30n = -2n^2 + 27n\n\Rightarrow -n^2 + 3n = 0\n\Rightarrow n(n - 
3) = 0\n$$\n\nSince $ n \neq 0 $, we get:\n$$\nn = 
3n$$\n\nSubstitute into $ S = -n^2 + 10n $:\n$$\nS = -9 + 30 = 
21\n$$\n\n---\n\n### Step 4: Compute the Current 
Mean\n\n$$\n\text{Current mean} = \frac{S}{n} = \frac{21}{3} = 
\boxed{7}\n$$\n\n---\n\n### Final Answer\n\n$$\nboxed{7}$$
\end{verbatim}
\end{tcolorbox}

\textbf{Ground Truth:}
\begin{tcolorbox}[
  breakable,
  enhanced,
  colback=gray!20,
  colframe=gray!20,
  boxrule=0.3pt,
  arc=2pt,
  sharp corners,
  valign=top
]
7
\end{tcolorbox}

\end{document}